\documentclass[acmsmall]{acmart}

\AtBeginDocument{%
  \providecommand\BibTeX{{%
    \normalfont B\kern-0.5em{\scshape i\kern-0.25em b}\kern-0.8em\TeX}}}

\usepackage{graphicx}
\usepackage{lipsum}
\usepackage{color}
\usepackage{framed}
\usepackage{colortbl}
\usepackage{graphicx}
\usepackage{xcolor}
\usepackage{cleveref}
\usepackage{siunitx}
\usepackage{amsmath}
\usepackage{subcaption}
\usepackage{gensymb}
\usepackage{verbatim}

\DeclareMathOperator{\area}{area}
\DeclareMathOperator{\closest}{closest}
\DeclareMathOperator{\dist}{dist}

\newcommand{\rebuttal}[1]{\textcolor{black}{#1}}
\newcommand{\crc}[1]{\textcolor{black}{#1}}

\begin{document}

\title{Jane Jacobs in the Sky: Predicting Urban Vitality with Open Satellite Data}

\author{Sanja \v{S}\'{c}epanovi\'{c}}
\email{sanja.scepanovic@nokia-bell-labs.com}
\author{Sagar Joglekar}
\email{sagar.joglekar@nokia-bell-labs.com}
\affiliation{%
  \institution{Nokia Bell Labs}
  \city{Cambridge}
  \country{U.K.}
}
\author{Stephen Law}
\email{slaw@turing.ac.uk}
\affiliation{%
  \institution{University College London \& The Alan Turing Institute} 
  \city{London}
  \country{U.K.}
}
\author{Daniele Quercia}
\affiliation{%
  \institution{Nokia Bell Labs}
  \city{Cambridge}
  \country{U.K.}
}

\renewcommand{\shortauthors}{\v{S}\'{c}epanovi\'{c}, et al.}

\begin{abstract}
The presence of people in an urban area throughout the day -- often called `urban vitality' -- is one of the qualities world-class cities aspire to the most, yet it is one of the hardest to achieve. Back in the 1970s, Jane Jacobs theorized urban vitality and found that there are four conditions required for the promotion of life in cities: diversity of land use, small block sizes, the mix of economic activities, and concentration of people. To build proxies for those four conditions and ultimately test Jane Jacobs's theory at scale, researchers have had to collect both private and public data from a variety of sources, and that took decades. Here we propose the use of one single source of data, which happens to be publicly available: Sentinel-2 satellite imagery. In particular, since the first two conditions (diversity of land use and small block sizes) are visible to the naked eye from satellite imagery, we tested whether we could automatically extract them with a state-of-the-art deep-learning framework and whether, in the end, the extracted features could predict vitality. In six Italian cities for which we had call data records, we found that our framework is able to explain \rebuttal{on average} 55\% of the variance in urban vitality extracted from those records.
\end{abstract}



\maketitle

\section{Introduction}

A livable city is a place full of life, and that life is created by city dwellers~\cite{montgomery1998making,lynch1984good}. In her 1961 book ``The Death and Life of Great American Cities'', writer and activist Jane Jacobs identified the four ``generators'' of urban vitality (i.e., pedestrian activity throughout the day): diversity of land use, small block sizes, a mix of economic activities, and the concentration of people. Without them, a city will die. With them, it will thrive.  Her ideas exerted a tremendous impact on the thinking of architects, developers, urban planners, and community activists. Despite their importance and popularity, for a long time, Jacobs's theories could not be tested at scale. Upon collection of both private and public data from a variety of sources, the hypothesized effectiveness of the four generators has been only recently tested in the city of Seoul, Korea \cite{sung2015operationalizing}, and across six Italian cities \cite{de2016death}. Yet, all these studies meticulously collected data from a variety of sources, and that took years (it took a decade in Seoul's case), considerably limiting research advancements. 

\begin{figure}
	\centering
	\includegraphics[width=.77\linewidth]{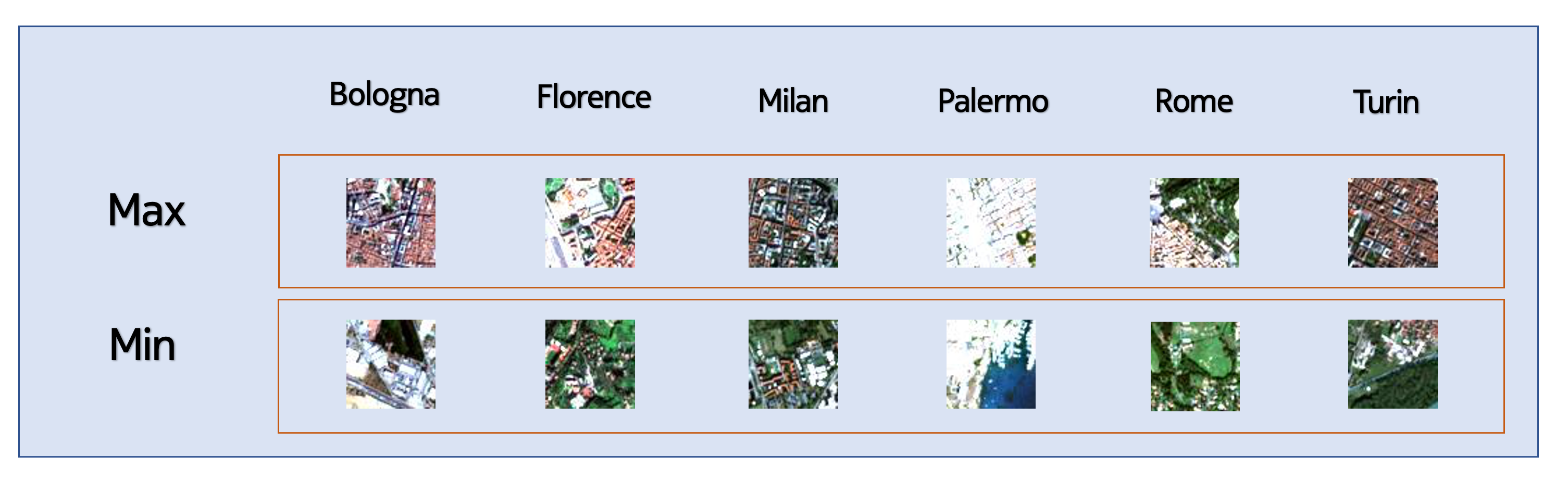}
	\caption{\rebuttal{Examples of satellite views for the areas with the highest and the lowest vitality levels in each of the six Italian cities. Despite having comparable vitality levels, the shown areas present very different urban forms.}\label{fig:labels}}
\end{figure}

To tackle the issue of data collection, we investigated whether it is possible to estimate some of the four generators of urban vitality (and vitality itself) from one single source: Sentinel-2 satellite imagery~\cite{torres2012}. Since two of the four generators---that is, diversity of land use, and small block sizes---are visible to the naked eye, we hypothesized that they could be identified from satellite imagery. We also studied whether vitality could be directly predicted from such imagery. \rebuttal{For both prediction tasks---predicting vitality generators first and then vitality indirectly from them, or predicting vitality directly---it is not obvious that neither of them could be performed upon satellite data. To see why, consider Figure \ref{fig:labels}: representative satellite views from the six Italian cities exhibit remarkably different visual features (such as colors and diversity of urban and natural forms) even if they come from areas of comparable vitality levels. }

We built a processing framework that consists of three modules, each of which: \emph{i)} extract small images from satellite imagery (which we call \emph{imagelets}); \emph{ii)} extract visual features from these imagelets with state-of-the-art deep learning methods; and \emph{iii)}  combine these features into district-level feature vectors. In so doing, we were able to make two main contributions:

\begin{itemize}
    \item[\textbf{C1}] We built two techniques based on a deep learning framework, which, upon publicly available Sentinel-2 satellite imagery, extract representative \emph{feature vectors for urban districts} (Section \ref{sec:methods}).  
    \item[\textbf{C2}] We showed that, from these \crc{district} feature vectors, we could predict proxies for \emph{land use} and \emph{small block size}, as well as predict \emph{urban vitality} directly (Section \ref{sec:results}).  In a 5-fold cross-validation experiment, we found that satellite features explain, on average across six Italian cities, more than $55\%$ of the variance \rebuttal{(in terms of the adjusted coefficient of determination, $R^2_{adj}$)} for urban vitality. Moreover, to ascertain \rebuttal{how well the model generalizes}, we performed a ``leave-one-city'' out validation (where the model is trained on a set of cities and tested on an unseen city). In this experiment, we found that the model can explain up to $50\%$ of variance for Milan and $61\%$ in Florence, suggesting the ability of our method to generalize to unseen cities. \rebuttal{However, we also found that this performance drops for some cities - up to $25\%$ for Rome. Upon a thorough investigation, we identified  the potential reasons for the performance drop in Rome, which include the city's specific natural, cultural, and historical contexts. Accordingly, we discuss recommendations for future work  (Section \ref{sec:discussion}). }
\end{itemize}

This work is at the intersection of two research areas---social computing and \crc{urban analytics}. The combination of the two makes it possible to use emerging sources of data such as satellite imagery to answer questions typical of the social sciences in a computational way, opening up new opportunities for future work (Section \ref{sec:discussion}). 

The project webpage is available at \url{https://goodcitylife.org/vitality}.

\section{Related Work}
Three main lines of research are related to this study. First, social computing research on quantifying urban activity.  Second, research that tested Jane Jacob's four generators of urban vitality at scale. Third, research on automatically inferring subtle urban properties from satellite imagery that are not necessarily visible to the naked eye (e.g., crime rates, property prices).

\subsection{Measuring Urban Activity from Social Media Data}
In the research community of social computing, researchers have worked on quantifying urban activity from different perspectives.  They have, for example, crowd-sourced people's perceptions of beauty, quietness, and happiness of urban locations~\cite{quercia2014shortest}, and, by then combining these crowd-sourcing results with geo-referenced social media content (e.g., Flickr photos), they built path recommender systems that suggest beautiful, quiet, or happy urban routes in a city ~\cite{peregrino2012mapping,van2011time,quercia2014shortest}.  \citet{le2015soho} used Foursquare check-ins to classify city neighborhoods in terms of prevalent activity types in them (e.g., dining, cultural, shopping). \citet{de2010automatic} identified the movements of individual tourists based on their picture uploading patterns and found that average time spent at a location and location popularity were two useful predictors of interesting routes.  

\rebuttal{Researchers used openly available social media data to study urban dynamics. For example, \citet{araujo2018visualizing} recently used the Facebook marketing API to demographically profile an urban area. \citet{redi2018spirit} predicted the ambiance of neighborhoods from the Flickr images in those neighborhoods. To augment deprivation indexes that have traditionally been derived from census data, and have been expensive to obtain, \citet{venerandi2015measuring} used Foursquare and OpenStreetMap data. The premise behind all this work is that user-generated content on location-based services translates into rich behavioural traces \cite{yoon2012social,de2010automatic,el2013photographer}}.  \rebuttal{More generally, in the \crc{social computing} community, a need to empower cities, neighborhoods, and local communities with new technologies has long been recognized  \cite{daly2015supporting}. In a similar vein, our work develops a framework that uses openly available data to quantify the key  aspect of urban vitality.}

\subsection{Verifying Empirically Jane Jacobs's Insights}
The  Household Travel Survey captures walking and driving activity in the whole city of Seoul and is conducted once every 5 years. \citet{sung2015operationalizing} used the 2010 survey to operationalize urban vitality and showed that the four generators of urban vitality did hold across dongs, the small administrative areas of Seoul.  In a similar setup, in six Italian cities, \citet{de2016death} have found that the four generators do apply to the Italian context as well. Instead of a costly multi-year survey, the researchers used mobile phone Internet density as a proxy for urban activity. Finally, beyond urban vitality, researchers have widely tapped into Jacobs's idea of \emph{natural surveillance} to build predictive models for the crime:  pedestrian activity acts as ``natural surveillance'', and that reduces crime rates~\cite{bogomolovmoves, bogomolov2014, traunmeller2014}. 

\subsection{Inferring Urban Variables from Satellite Imagery}
Satellite images capture an overall structure of an area, which encodes different urban processes, as recent studies \crc{on} satellite images have shown.
\rebuttal{For example,} \citet{han2019lightweight} developed a deep learning architecture to map high-resolution satellite images across country districts into fixed-size district-level feature vectors. \crc{The authors then} showed that such features can be used to predict socio-economic properties of districts, such as population density, population age distribution, household count and size, and income per capita.
\citet{albert2017using} studied the patterns of land use in urban neighborhoods using high-resolution satellite imagery from Google Maps. To label the imagery, \crc{the authors} resorted to the Urban Atlas,
which provides land classification into 20 land use classes across 300 European cities. For a given location, \crc{they trained ResNet \cite{he2016deep} and VGG-16 \cite{Simonyan2014VeryDC}, two deep convolutional neural networks, }to predict the most likely class for the $224 \times 224$ images. The achieved classification accuracy was between 0.7 and 0.8. Moreover, \crc{\citeauthor{albert2017using}} showed that the models can be trained on the imagery from one city to predict the classes in a completely different city. \citet{law2019take} studied whether satellite\crc{/aerial} imagery can enhance the prediction of housing prices. \crc{The authors} obtained the high-resolution satellite\crc{/aerial} imagery from Bing around the properties of interest and showed that the standard hedonic price approach for predicting housing prices can be improved by incorporating features extracted from the satellite imagery in a semi-interpretable manner. 

\citet{albert2018dark} studied the distribution of key urban macroeconomic variables -- population, luminosity (a proxy for energy access and use), and building density -- from satellite data. Across all the cities with at least 10K inhabitants around the world, \crc{the authors} demonstrated the strong link between the spatial distribution of lighting levels (which was previously shown to be a proxy for energy access and wealth levels \cite{jean2016combining}) and concentration of population. 
\rebuttal{The work by \citet{albert2018dark} is similar to ours, in that, the authors studied population density and built infrastructure, both of which are proxies for urban vitality. However, \crc{the} work \crc{by \citeauthor{albert2018dark}} differs from ours in two main ways: \emph{i)} \crc{there is no operationalization of}  the theoretical concept of urban vitality as this was not the focus; and \emph{ii)} \crc{the authors studied} 25K cities at a coarse-level  (with TerraSar and LandScan data at a spatial resolution of $750m-1km/px$), while \crc{in this present work, we were} able to work at a much finer-level of urban detail (with Sentinel-2 data at a spatial resolution of $10m/px$).}

\rebuttal{ \citet{wang2018urban} studied whether the {commercial activeness} of an urban area, proxied with the number of online reviews for places in the area, can be predicted from satellite data (and street views). \crc{The work} \crc{by  \citeauthor{wang2018urban}} is similar to ours, in that,  a pipeline \crc{was used} that extract features from satellite images and feeds those features into a regression method. However, \crc{there are two main differences compared to our present study}: \emph{i)} \crc{there is no investigation} \crc{of} urban vitality but only one of its components, i.e., commercial activeness; and \emph{ii)} \crc{\citeauthor{wang2018urban} work} used \emph{commercial} satellite imagery, compared to the \textit{openly available Sentinel-2 imagery}  \crc{that we} used. \crc{The commercial nature of the data might create hurdles in terms of inclusivity in the social computing community, especially for the developing world.} }

\rebuttal{ In summary, though previous work used satellite data to measure population density, commercial activeness, and other variables that influence urban vitality, no study directly focused on vitality. Furthermore, the majority of previous work used commercial satellite images such as those from Bing and Google Earth, instead of openly available Sentinel images, as we did. }\crc{By basing this work on openly available satellite data, such as Sentinel, we show that we can systematically derive insights about the built environment (such as vitality) with a performance at par with methods based on other access controlled and expensive datasets. This is notwithstanding the differences in spatial resolution between Sentinel and other datasets. We foresee that this could further empower the community to perform follow-up studies, particularly around the developing parts of the world.}

\section{Data and methods}\label{sec:methods}
This section describes how each city's district was encoded into a \emph{district feature vector}, which was then associated with the district's urban vitality value.  First, we created \emph{imagelet feature vectors} by  extracting small image pieces (imagelets) from satellite imagery and parsing them using deep learning feature extractors (Section \ref{sec:features}). Second, we computed the district's urban vitality by combing  a variety of data sources, including Urban Atlas, Open Street Map, and mobile phone data (Section \ref{sec:label}). Finally, we created the district's feature vector by combining the \emph{imagelet feature vectors} corresponding to the imagelets composing the district, and associated the resulting  \emph{district feature vector} with the  previously computed district's urban vitality value (Section \ref{sec:combined}).

 


\begin{figure*}
	\centering
	\includegraphics[width=.92\textwidth]{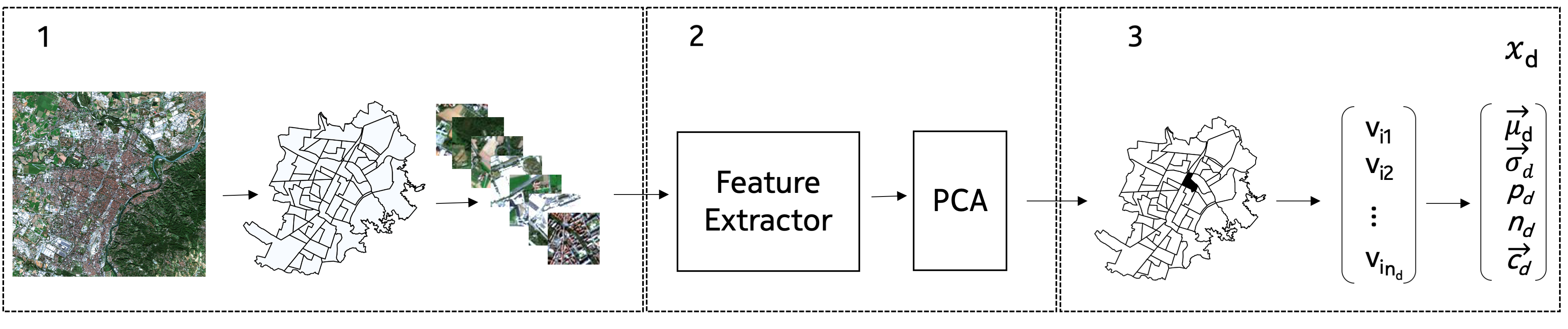}
	\caption{Our framework for district feature extraction from satellite imagery. In step 1, the entire Sentinel-2 image of the city is split into imagelets (see Figure \ref{fig:preprocessing_milano}); in step 2, features are extracted from each imagelet using a deep learning feature extractor (see Figures \ref{fig:network_struc_pretrained} and \ref{fig:network_struc_autoencoder}), which is then followed by a PCA; and, in step 3, the imagelets $i$ are grouped by their corresponding districts (Figure \ref{fig:TorinoLabels}) and their feature vectors ($v_{i}$) processed to finally compute the representative district feature vectors ($x_d$).  \label{fig:framework}}
\end{figure*}

\subsection{{Creating Imagelet Feature Vectors from Satellite Images}\label{sec:features}}
In 2014, the European Space Agency (ESA) has launched the first of the Sentinel satellites as part of the Copernicus\footnote{https://www.copernicus.eu/en/about-copernicus/copernicus-brief} program \cite{torres2012}, which aims at democratizing  access to Earth Observation (EO) data. Thanks to this multi-billion investment, we can today freely access accurate and timely land satellite data from the radar (Sentinel-1, Sentinel-3), altimeter (Sentinel-3), and optical (Sentinel-2) sensors. Other satellites from the Sentinel series provide atmospheric and oceanic data. Goals behind programs such as Copernicus are quite ambitious: they include  managing the environment, helping mitigate the effects of climate change, monitoring agriculture, and supporting urban development \cite{showstack2014sentinel}, to just name a few. 

\subsubsection{Sentinel-2 Imagery}
For this study, we used Sentinel-2 optical imagery, which monitors the land surface conditions and serves for producing land-cover and land-change detection maps  \cite{drusch2012sentinel}. The optical sensors (Multi-Spectral Instrument \cite{chorvalli2010design}) sense 13 spectral bands (B1-B13) ranging in spatial resolution from 10$m$ to 60$m$. Out of the several processing levels in which the Sentinel-2 imagery is distributed, Level-1C and Level-2A are freely available to public. The Level-2A is the higher processing level and it includes atmospheric correction \cite{main2017sen2cor} on top of Level-1C products. Additionally, the imagery in this product is orthorectified (projected to the geodetic coordinates) and generated with an equal spatial resolution of 10$m$ for the three so-called True Color Image (TCI) bands (i.e., B4, B3, B2). The TCI bands combination gives natural color representation of Sentinel-2 data \cite{gatti2013sentinel} and is widely used in land cover studies \cite{sovdat2019natural}. Given its richness and that all these necessary pre-processing steps are already performed on it, we selected the Level-2A product for this study. We downloaded the Sentinel-2 Level-2A products for selected six Italian cities from 2018 (i.e., the earliest available, since this product became operational in 2018). All the images were re-projected to the \textit{WGS 84 / UTM zone 32N} geodetic coordinate reference system (CRS), also sometimes denoted as \textit{EPSG:32632}. This CRS is suitable for use in between 6\degree E and 12\degree E, northern hemisphere and between equator and 84\degree N -- the area that includes Italy. 

\subsubsection{Pre-processing and Creating Imagelets}
\begin{figure}
	\centering
	\includegraphics[width=.777\textwidth]{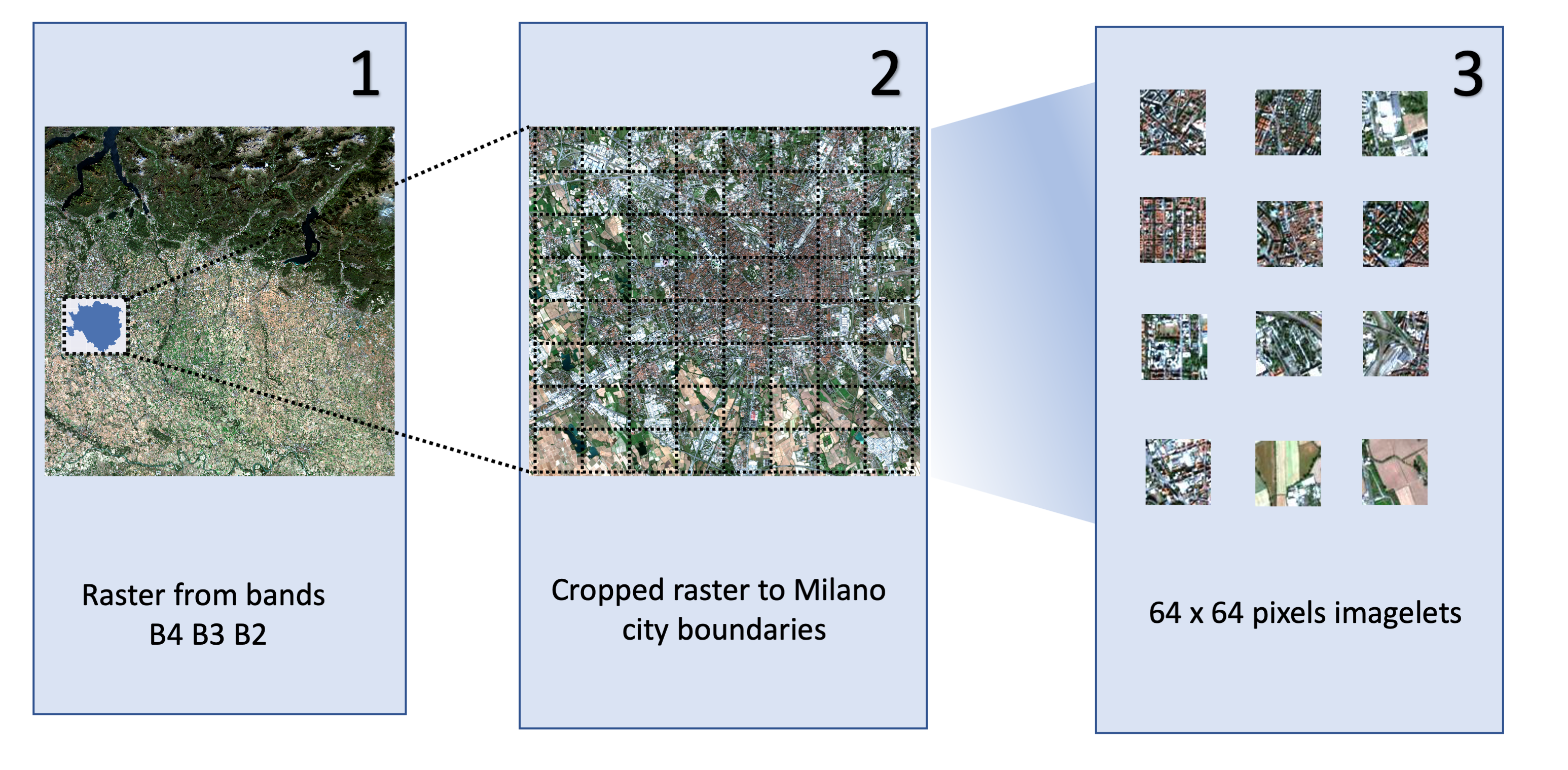}
	\caption{Process of creating training imagelets from the source Sentinel-2 imagery for the city of Milan.\label{fig:preprocessing_milano}}
\end{figure}

To create training data, for each city, we processed the Sentinel-2 imagery in three steps (Figure \ref{fig:preprocessing_milano}):
\begin{enumerate}
	\item created geo-referenced raster image from the TCI bands (B4, B3, B2);
	\item cropped the image based on the city boundaries (derived from an official shapefile); and
	\item split the cropped image into imagelets of size $64 \times 64$ pixels.
\end{enumerate}
This imagelet size was selected so that the resulting imagelets are not too large compared to a district's size (Table \ref{table:stats_cities}) and not too small in terms of their final resolution, but they are also enough large to be successfully processed by our deep learning framework (Section \ref{sec:DLextractors}). The pre-processing resulted in $9{,}115$ imagelets in total across the six cities.

\subsubsection{Deep Learning Feature Extractors\label{sec:DLextractors}}
\begin{figure*}[t]
	\centering 
	\begin{minipage}[b][][b]{.49\textwidth}
		\centering 
		\includegraphics[width=0.99\textwidth] {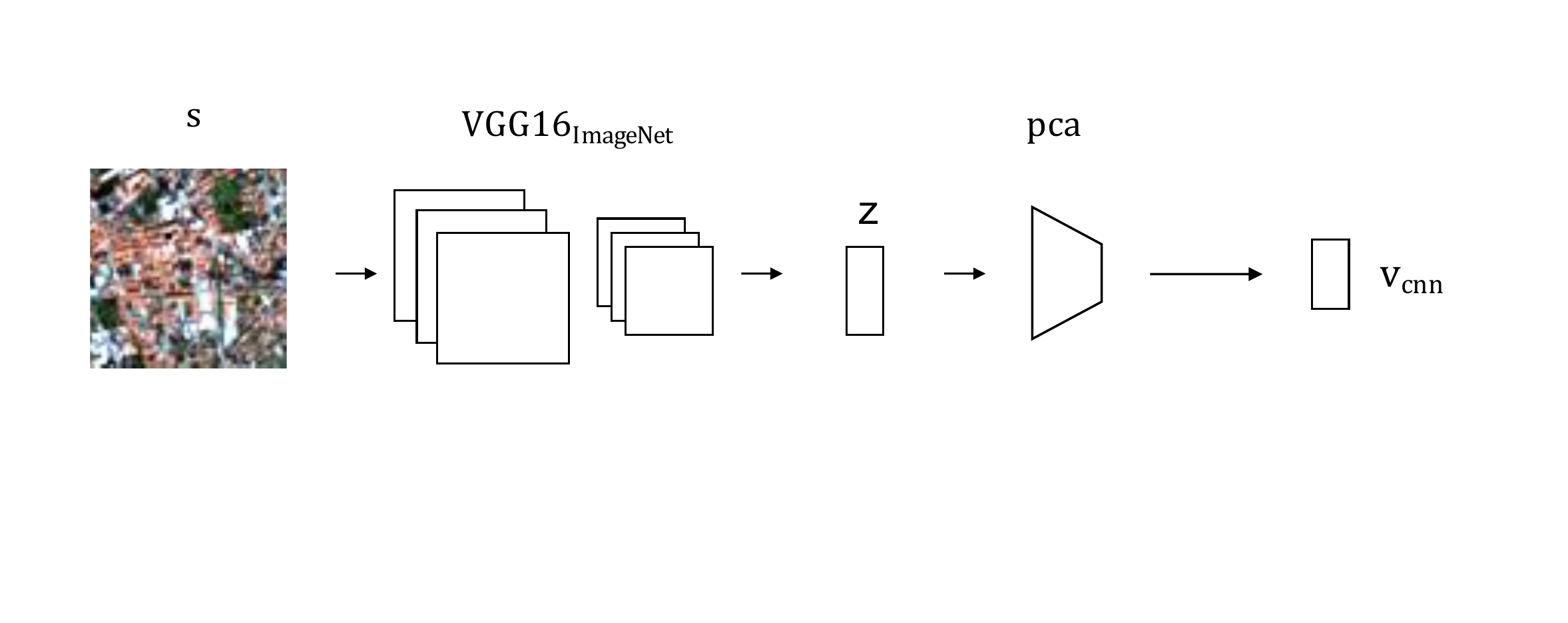}
		\captionof{figure}{Schematic of our pre-trained $CNN$ feature extractor. 
			{We extracted a set of features $z$ from satellite image using a pretrained $VGG16_{imagenet}$ \cite{simonyan2014deep}. We then summarised these features into a set of components  $v^{CNN}$ with a $PCA$.}}
		\label{fig:network_struc_pretrained}
	\end{minipage}\hfill
	\begin{minipage}[b][][b]{.49\textwidth}
		\centering 
		\includegraphics[width=0.99\textwidth] {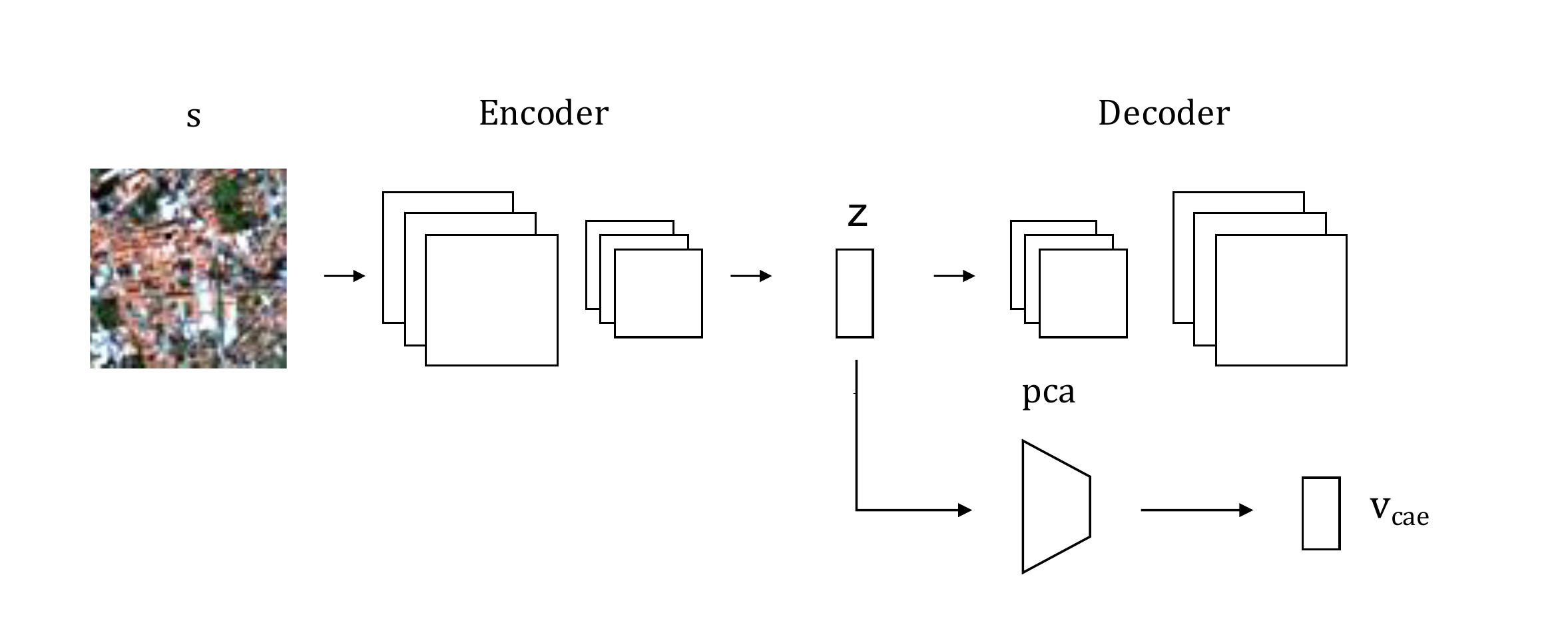}
		\vspace{0.5em}
		\captionof{figure}{Schematic of our Convolutional Auto Encoder ($CAE$) feature extractor. 
			{We trained a $CAE$, where the encoder compresses the satellite image and a decoder reconstructs it. After training, we extracted a set of features $z$ from the encoder and summarise it into a set of components $v^{CAE}$ with a $PCA$.}} 
		\label{fig:network_struc_autoencoder}
	\end{minipage}
\end{figure*}

To extract visual features from the satellite imagelets, we applied a pipeline consisting of a standard convolutional neural network ($CNN$)\crc{~\cite{krizhevsky2017imagenet}} architecture followed by a principal component analysis ($PCA$). Given an image $i \in I$, where $I$ is the set of imagelets, our goal was to retrieve a vector of uncorrelated and ordered visual components $v_i \in V$. To generate such a vector we used two techniques: the first was based on a pre-trained feature extractor (Figure~\ref{fig:network_struc_pretrained}), and the second consisted of training an unsupervised convolutional autoencoder (CAE) feature extractor (Figure~\ref{fig:network_struc_autoencoder}).

\paragraph{Pre-trained CNN Feature Extractor.} 

We adopted the $VGG16$ \cite{simonyan2014deep} $CNN$ architecture (Figure \ref{fig:network_struc_pretrained}) (weights pre-trained on ImageNet \cite{russakovsky2014imagenet}): given an imagelet $i$, we extracted the output of the last but one fully connected layer ($4{,}096$ features) ($z_i$) and summarised these features $z_i$ into a sparse set of uncorrelated components $v_i^{CNN}$ using a $PCA$.

\paragraph{Convolutional \crc{Autoencoder} (CAE) Feature Extractor.} 
{Since $VGG16$ architectures were not trained on ``overhead view'' satellite images, we chose an additional architecture which we could train on our imagelets. This architecture uses a convolutional autoencoder ($CAE$) \cite{CAE, AE} (Figure \ref{fig:network_struc_autoencoder}), followed by  $PCA$ \cite{Law2019b}. $CAE$ is an unsupervised approach which comprises of two parameterized functions, a deterministic encoder $f_w( \cdot )$, and a deterministic decoder $G_u( \cdot )$. Convolutional layers can be stacked sequentially where the encoding layers reduce the dimension of the image $i$ to a latent embedding $z_i$ while the decoding layers expand the dimensions back to its reconstruction $i^\prime$. The sequential architecture can be seen in Figure \ref{fig:network_struc_autoencoder}.
Following \cite{CAE}, the parameters of the encoder $z_i=F_w(i)$ and the decoder $i^\prime = G_u(z_i)$ are updated by minimizing the reconstruction losses between $i$ and the reconstructed version $i^\prime=G_{u}(F_w(i))$:}


\begin{equation}\label{eq:Loss}
\mathcal{L}_{REC}= \frac{1}{n_{batch}} \sum_{j = 0}^{j=n_{batch}} (i_j-G_{u}(F_w(i_j)))^2,
\end{equation}
where $n_{batch}$ is the number of images in a batch, and $\mathcal{L}_{REC}$ is the reconstruction loss.

The convolutional autoencoder was trained with the satellite images using the ADAM\crc{~\cite{kingma2017adam}} optimiser with a learning rate of 0.001 for 500 epochs with a batch size of 128. After extensive testing, we selected an embedding dimension of 512 since it gave us the best trade-off between compression and a lower reconstruction loss, where the loss stabilizes at 0.01 (Table \ref{table:CAE} reports the implementation details of the architecture we ended up using).
After training, we extracted a lower dimension embedding from the satellite imagelets using the trained encoder, which is then summarized into uncorrelated components $v_i^{CAE}$ using $PCA$.  

For both, $VGG16$ and CAE, we experimented with $n_{comp}=12,...,64$ PCA components in our evaluation (Section \ref{sec:results}), and found that using $n_{comp}=16$ components yielded the best results when using all the data from our six cities. In the rest of the paper, for simplicity, we denote imagelet feature vectors with $v_i$, while we experimented with both $v_i^{CNN}$ and $v_i^{CAE}$.



\begin{table}[t!]
	\caption{Number of districts ($N_c$), their average size (in \si{\square\km}) and population, and number of imagelets acquired per city.  \label{table:stats_cities}}
	\footnotesize
	\begin{tabular}{lllll}
		City & \#Districts ($N_c$) & Mean Area (in \si{\square\km}) & Mean Population & \#Imagelets  \\
		\hline
		Milan & 85 & 1.72 & 14,551 & 328 \\
		Bologna & 23 & 3.34 & 15,918 & 170  \\
		Florence & 21 & 2.89 &  16,633 &  135  \\
		Palermo & 43 & 2.01 & 15,075 & 193  \\
		Turin & 56 & 2.00 & 15,543 & 252  \\
		Rome & 146 & 3.24 & 17,123 &  1068 \\
		\hline
	\end{tabular}
\end{table}

\begin{table}[t!]
	\caption{$CAE$ Architecture \crc{with explicitly enumerated stack of \emph{encoder} and \emph{decoder} layers}  \label{table:CAE}}
	\footnotesize
	\begin{tabular}{llll}
		LAYER & RESAMPLE & NORM & FILTER SIZE  \\
		\hline
		Image & - & - & -\\
		\hline
		Conv & MaxPool & BatchNorm & 5$\times$5$\times$16  \\
		Conv & MaxPool & BatchNorm & 7$\times$7$\times$32  \\
		Conv & MaxPool & BatchNorm &  7$\times$7$\times$32   \\
		Conv & - & BatchNorm & 9$\times$9$\times$32  \\
		Conv & - & BatchNorm & 9$\times$9$\times$32 \\
		\hline
        Z & & & 512$\times$1 \\
        \hline
		ConvTranspose & Upsample & - & 9$\times$9$\times$32  \\
		ConvTranspose & Upsample & - & 9$\times$9$\times$32  \\
		ConvTranspose & Upsample & - &  7$\times$7$\times$32 \\
		ConvTranspose & Upsample & - & 7$\times$7$\times$32  \\
		ConvTranspose & Upsample & - & 5$\times$5$\times$16  \\
		\hline
		Recon & - & - & -\\
		\hline
	\end{tabular}
\end{table}

\subsection{Computing Six Vitality Proxies and  Urban Vitality for a District\label{sec:label}} 
We built upon previous work by \citet{de2016death}. The authors empirically validated Jane Jacobs theory across six cities in Italy. They showed that the four generators of urban vitality defined by Jacobs, i.e., \emph{land use, small blocks, diversity of economic activity}, and \emph{concentration of people}, indeed predicted urban vitality. 

From our satellite imagery, we can estimate neither diversity of economic activity (it is not a property of the urban form) nor concentration of people (as our satellite images are open data but medium-resolution). By contrast, land use and small blocks are the two vitality generators out of the four that could be potentially visible from satellites.  To test the extent to which they are so, we operationalized them with \emph{six vitality proxies}  (Table \ref{table:selected_JJ_vars}) computed for the six cities (Milan, Florence, Bologna, Turin, Palermo, and Rome).  



\begin{table}[t!]
\caption{The six vitality proxies plus urban vitality itself together with their distribution, mean ($\mu$), and standard deviation ($\sigma$). \label{table:selected_JJ_vars}}
\footnotesize
\begin{tabular}{llllllll}
\hline
\textbf{Land Use} &  & $\mu$ & $\sigma$ & \textbf{Small Blocks} & & $\mu$ & $\sigma$ \\

 Land use mix & \includegraphics[width=0.17\columnwidth]{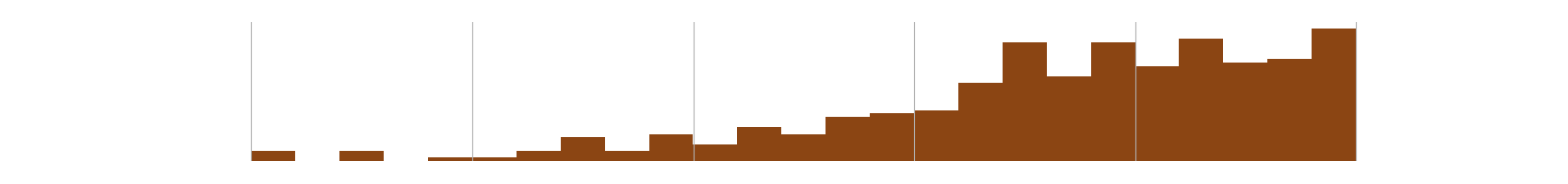} & 0.733 & 0.201 & Block size & \includegraphics[width=0.17\columnwidth]{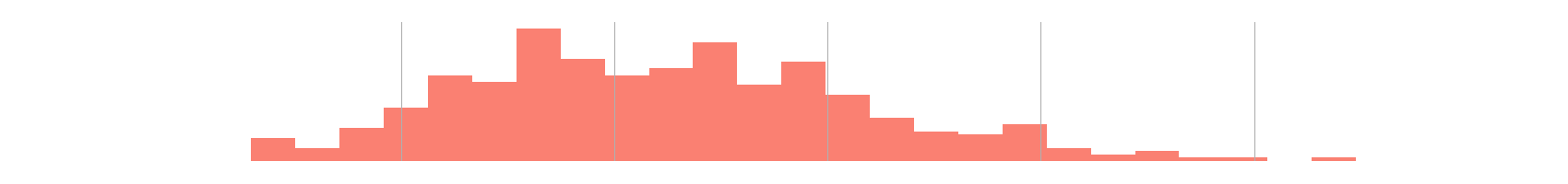} & 9.618  & 0.459  \\
Building height & \includegraphics[width=0.17\columnwidth]{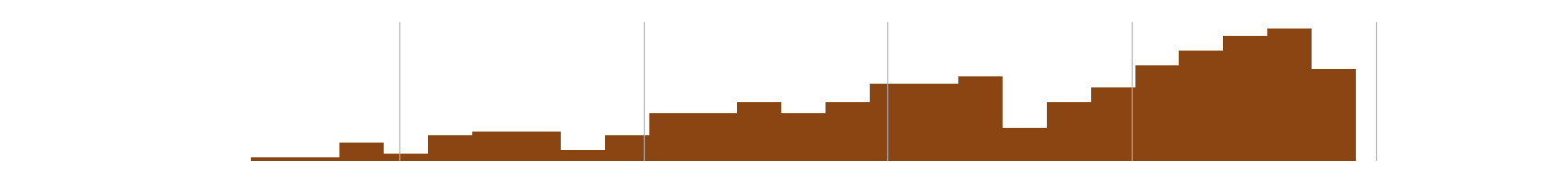} & 0.689 & 0.216& Intersection density  & \includegraphics[width=0.13\columnwidth]{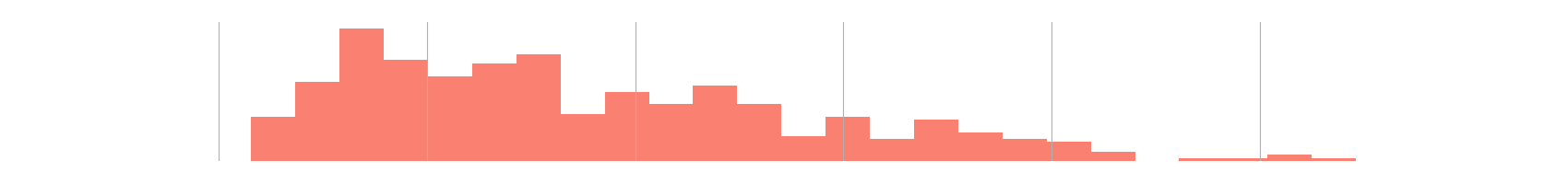} & $10^{-4}$  & $10^{-4}$  \\
 Small parks & \includegraphics[width=0.17\columnwidth]{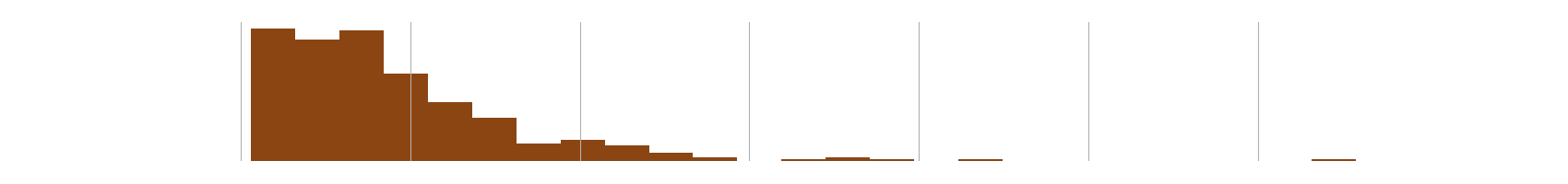} & 0.004 & 0.003 & Anisotropicity &  \includegraphics[width=0.17\columnwidth]{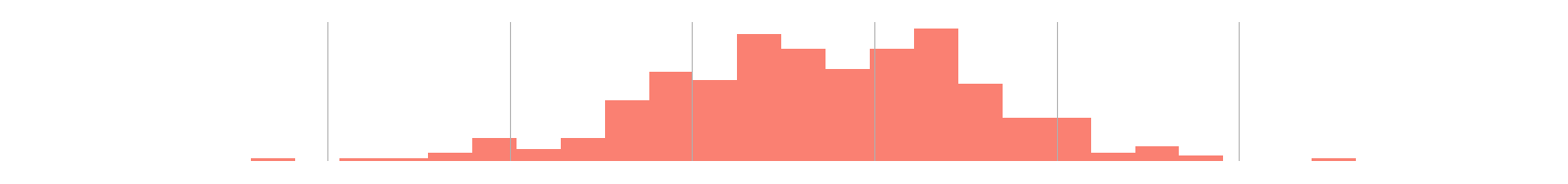} & 0.385 & 0.042\\
   &   \\
  \textbf{Vitality}  &  & $\mu$ & $\sigma$ &  & &  &   \\
   Activity density &  \includegraphics[width=0.17\columnwidth]{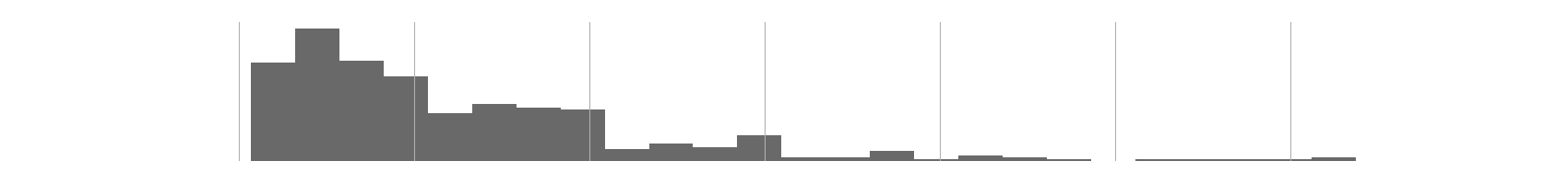}  & 0.006 & 0.005 &  &  &  &   \\
 
 \hline
\end{tabular}
\end{table}


Before describing these six proxies, we have to define and motivate our choice for spatial unit of analysis. We took \textit{district} (or \textit{area di censimento}) as the unit of analysis, as was done by previous work \cite{de2016death}. Districts are census areas, which consist of neighboring blocks (i.e., sections delimited by street segments) grouped based on socio-economic conditions. In the Italian context, districts are comparable in terms of population (having between 13K and 18K inhabitants) and size (covering an average area of $2.47 km^2$) (Table \ref{table:stats_cities}). The average population density of a district for the 6 cities in our dataset is 10K per $km^2$. \rebuttal{These parameters correspond to those of the spatial areas that Jacobs discussed in her book \cite{jacobs2016death}.}


\subsubsection{Land Use.} 
Jacobs suggested that a mix of primary uses in a district promotes vitality~\cite{jacobs2016death}. Examples of primary use categories are residential buildings, office spaces, industrial, entertainment, education, recreation, and cultural facilities. One of the most comprehensive datasets on the urban land use categories is  the  Urban Atlas\footnote{\url{http://www.eea.europa.eu/data-and-maps/data/urban-atlas}} produced for most of the cities in Europe including our test cities. It provides information on 20 land use classes (such as green urban areas, sports and leisure facilities, and urban fabric). The Urban Atlas is built from high resolution satellite imagery in combination with ancillary data, such as Google Earth,  Open Street Map, and manually-collected in-situ data.

\paragraph{Land use mix.} From the Urban Atlas, we first computed  \textit{land use mix} in each district. According to \citet{manaugh2013mixed}, land use mix can be calculated as the entropy of three main categories of urban land uses: i) \emph{residential}, ii) \emph{commercial, industrial}, \emph{institutional} and \emph{governmental}, and iii) \emph{recreational, parks}, and \emph{water}. We used the formula:
\begin{equation}
\text{land use mix}_d = - \sum_{j=1}^{n_{LUC}} \frac{P_{d,j} \log(P_{d,j})}{\log(n_{LUC})}
\label{eq:LUM}
\end{equation}
where $P_{d,j}$ is the percentage of area with land use $j$ in district $d$, and $n_{LUC}=3$ is the number of land use categories.

\paragraph{Small parks.}  By attracting people to spend time or just walk there, small parks promote vitality \cite{jacobs2016death,banchiero2020neighbourhood}. For each district $d$, we calculated the average distance of all its blocks from the nearest small park (with the area $< 1$ \si{\square\km}):
\begin{equation}
\text{small parks}_d = (\frac{1}{|B_d|} \sum_{j \in B_d} \dist(j, \closest(j, SM)))^{-1}
\label{eq:CSM}
\end{equation}
where $B_d$ is the set of blocks in district $d$, $\closest(j, Y)$ is a function that finds the geographically closest element in set $Y$ from block $j$'s centroid, $SM$ is the set of small parks, and $\dist(a,b)$ is the geographic distance between two elements' centroids $a$ and $b$.

\paragraph{\rebuttal{Building height.} } Jacobs posited that lower building height encourages the opening of restaurants, stores, and other services, which, in turn, all promote pedestrian activity \cite{jacobs2016death}.  \citet{sung2015operationalizing} used the average number of floors per building in a district as a proxy for \emph{building height}, which was computed as:
\begin{equation}
\text{building height}_d = \frac{\sum_{hc} b_{hc,d}\cdot f_{hc}}{\sum_{hc} b_{hc,d}}
\label{eq:HT}
\end{equation}
where $b_{hc,d}$ is the number of buildings that are in height category $hc$ in district $d$, and $f_{hc}$ is the number of floors corresponding to height  category $hc$.  To collate data on building floors, we used the Italian National Institute for Statistics (ISTAT)\footnote{\url{http://www.istat.it/it/archivio/104317}}.

\subsubsection{Small Blocks.} 
Another of our six vitality proxies is small blocks, which have been found to  increase walkability and opportunities of cross-use.

\paragraph{Block sizes.} A district's \emph{block size} variable can be defined as the average size of all the blocks $B_d$ within the district $d$:
\begin{equation}
{\text{block sizes}}_d = \frac{1}{|B_d|} \sum_{j \in B_d} \area_j
\label{eq:BA}
\end{equation}

\paragraph{Intersection density.} Another variable in the category of small blocks contributing to urban vitality by increasing random contacts is \textit{intersection density}:
\begin{equation}
\text{intersection density}_d = \frac{|\text{intersections}_d|}{\area_d}
\label{eq:ID}
\end{equation}

\paragraph{Anisotropicity.}  
\rebuttal{Anisotropy refers to the \crc{irregularity} of a geometric shape concerning its orientation and spacing. A block can be of relatively small size, but, if it exhibits an anisotropic shape (i.e., one of its sides is  considerably longer than the other), then that decreases the chance of being in contact with one of its sides.} We assigned \textit{anisotropicity} to each district $d$  as the average anisotropicity~\cite{louf2014typology} of its blocks $B_d$:
\begin{equation}
\label{eq:phi}
\text{anisotropicity}_d = \frac{1}{|B_d|} \sum_{j \in B_d} \Phi_j
\end{equation}
where $\Phi_j$ is the ratio between the area of the block $j$ and the area of its circumscribed circle $\mathcal{C}_j$:
\begin{equation}
\Phi_j=\frac{\area_j}{\area_{\mathcal{C}_j}}.
\end{equation}

\subsubsection{Urban Vitality.} 
We used mobile Internet activity as a proxy for urban vitality. Unlike mobile phone data on calls and SMS activity, this type of data senses peoples' presence even when they are not actively using their phones, enabling better tracking of their mobility in an area. Telecom Italia Mobile, i.e., the largest mobile operator in Italy ($34\%$ of total mobile phone user base), provided its data from February to October 2014. 

\paragraph{Activity density.}  To estimate the number of Internet connections that fell into each district, we represented the space with a set of Voronoi polygons \cite{du1999centroidal} based on the radio stations' positions. To estimate the Internet activity in each district $d$ at time $t$, we calculated the number of Internet connections over all polygons $p$'s that fell into district $d$: 
\rebuttal{
\begin{equation}
S_{d} = \sum_{v } R_{p} \frac{A_{p \cap d}}{A_p - A_{p \cap W}}
\label{eq:activity} 
\end{equation}
where $p$ is a polygon, and $R_{p}$ is the number of Internet connections in $p$ throughout a typical business day. The count of Internet connections is weighted by $\frac{A_{p \cap d}}{A_p - A_{p \cap W}}$, which is the proportion  $\frac{A_{p \cap d}}{A_p }$ of $p$'s area that falls into district $d$ ($A_{p \cap d}$ is $p$'s area that falls into district $d$, and $A_p$ is  $p$'s total area). From $p$'s total area we removed sea areas denoted by $W$ (i.e., we removed $A_{p \cap W}$). }

\subsection{Combining District Feature Vectors with Urban Vitality\label{sec:combined}} 
\begin{figure}
\centering
\includegraphics[width=.92\linewidth]{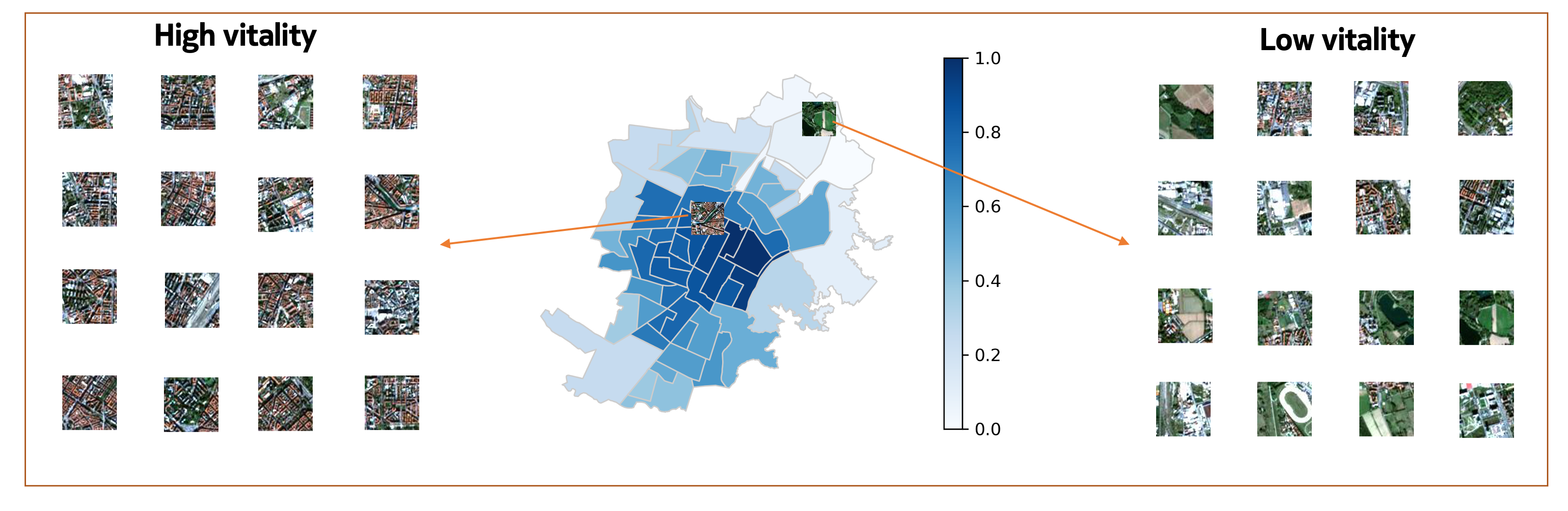}
\caption{Our approach of associating imagelets with districts and their vitality levels (which are computed from mobile phone activity density and are our labels) in the city of Turin. A darker color denotes higher values of urban vitality. The two imagelets shown on the map will be associated with different vitality levels (i.e., the imagelet in the city center with high vitality, while the one on the outskirts with low). Examples of imagelets associated with high versus low vitality levels are also shown. The high-vitality imagelets are characterized by a denser urban fabric, smaller blocks, and more brown-reddish colors (likely representing the rooftops of low-rise buildings). On the other hand, the low-vitality imagelets feature large homogeneous areas (e.g., stadium, larger parks, and fields), a low-density urban fabric, and white colors (likely representing industrial buildings).
\label{fig:TorinoLabels}}
\end{figure}

Having prepared imagelet-level features and district-level labels, we combined the two datasets as follows. First, we assigned imagelets to districts with the following procedure. If an imagelet was overlapping with several districts, we assigned  it to the district with which it has the largest overlap. An illustration for assigning imagelets in Turin is shown in Figure \ref{fig:TorinoLabels}. Out of the $9{,}115$ original imagelets, most of them fell outside of the city's boundaries. With this procedure, we were able to assign a different percent of imagelets for each city, in total $2{,}146$ imagelets (the breakdown per city is shown in Table \ref{table:stats_cities}). \rebuttal{Notice, however, that we still used all the $9$K imagelets to train the autoencoder feature extractor $CAE$ (described in Section \ref{sec:features}).} These imagelets constituted the basis for our train and test set. 

Second, using the imagelet features, for each city and each of its districts, we derived a representative district feature vector in a way similar to the previous work \cite{han2019lightweight}. For each district $d \in 1,...,N$, we took feature vectors $v_{i_1}, v_{i_2}, ... v_{i_{n_{d}}} \in R^{n_{comp}}$ (where $n_{comp}$ is the number of PCA components discussed in Section \ref{sec:features}) of its imagelets $i_1, i_2, .., i_{n_d}$, and aggregated them into a single vector $x_d = (\vec{\mu_d}, \vec{\sigma_d}, p_d, n_d, \vec{c_d)}$, where $\vec{\mu_d}$ is the vector of size $n_{comp}$ obtained by taking the mean across the $n_{comp}$ components of the district imagelet feature vectors, $\vec{\sigma_d}$ is the vector also of size $n_{comp}$ obtained in a similar way by taking the standard deviation across the components of imagelet vectors, $p_d$ is a number obtained by taking the the average Pearson's correlation between pairs of imagelet vectors, $n_d$ is the total number of imagelets in $d$, and $\vec{c_d}=({c_d^{lon}, c_d^{lat}})$ is a vector of size two representing the district centroid (i.e., the lon/lat of the centroid point). The first four groups of elements capture and describe district satellite data by central tendency ($\vec{\mu_d}$), dispersion ($\vec{\sigma_d}$), association ($p_d$), and size ($n_d$). Compared to \cite{han2019lightweight}, we additionally enriched our visual feature set with geolocation features, i.e., district centroids ($\vec{c_d}$) as previous research has shown that their inclusion encodes spatial positioning of the imagelets, \rebuttal{and improves performance} \cite{Tang_2015,Chu_2019}. Hence, if we denote as $x_d^{k}$ the $k^{th}$ element of the feature vector $x_d$, then: 
\begin{description}
	\item[The first $n_{comp}$ elements] are the average values of the $1^{st}$, $2^{nd}$, \ldots, $n_{comp}$-$th$  component (respectively) of all the district's imagelet vectors, 
	
	\item[The elements from position $(n_{comp}+1)$ to position $(2 \cdot n_{comp})$] are the standard deviation values of the $1^{st}$, $2^{nd}$, \ldots, $n_{comp}$-$th$ component of all the district's imagelet vectors,
	\item[The element at position $(2n_{comp}+1)$] is the Pearson's correlation of the imagelet vectors,
	\item[The element at position $(2n_{comp}+2)$] is the count of imagelets in the district, and
	\item[The last two elements] are the longitude and latitude coordinates of the district's centroid.
\end{description}
 Our way of defining the district feature vectors has the nice property that they are of equal length (i.e., $x_d \in R^m$, where $m=2n_{comp}+1+1+2$), independently of the district sizes or the number of imagelets.

Last, for each of the collected six proxies (Table \ref{table:selected_JJ_vars}), we took the corresponding value $y_d$ for the district $d$ and associated this value with the feature vector $x_d$ of the district. This combined dataset constituted our training and test set.



\section{Evaluation}\label{sec:results}
Having prepared the training and test data, we set up our two main evaluation experiments. The main goals of these experiments were to test whether we could estimate two classes of quantities  from satellite imagery:  \emph{1)}  the six proxies in Table~\ref{table:selected_JJ_vars}  (Section~\ref{sec:predict_JJ_vars}), and \emph{2)} urban vitality itself (Section~\ref{sec:predict_vitality}).




\subsection{Setup}\label{sec:setup}
Using the produced district feature vectors, for each of the six proxies in Table \ref{table:selected_JJ_vars}, we created a regression task whose dependent variable is initially each of the six vitality proxies. The distributions of different variables, including activity density used as a proxy for urban vitality, are shown in Table \ref{table:selected_JJ_vars}. Since activity density and the variables in the \textit{land use} category were skewed, we log-transformed them using the natural logarithm. Additionally, since the scales of the variables varied considerably, we standardized and normalized them before running regression models so that we could later compare their relative importance.

\subsection{Regression methods}
Given the district feature vectors $x_{d} \in R^m$ (calculated as in Section \ref{sec:combined}) and their corresponding labels $ y_{d} \in R$ (calculated as in Section \ref{sec:label}), we aimed at predicting $\hat{y_{d}}$, 
using linear regression models of the form
\begin{equation}\label{eq:regression}
\hat{y_{d}} = w_0 + w_1 x_{d}^1 + ... + w_m x_{d}^m \text{,} \quad  d = 1,...,N,
\end{equation}
where the coefficients $w_j$ are learned by the model, $m$ is district feature vector size, and $N$ is the total number of districts.

In addition to \textbf{Ordinary Least Squares regression}, which minimizes $||y_d - X_dw||^2_2,$
where $X_d$ denotes the matrix of all district vectors $x_{d}$; we also experimented with \textbf{ElasticNet regression}, which minimizes $
\frac{1}{2N}  ||y_d - X_dw||^2_2
+ \alpha  \rho ||w||_1
+ \frac{\alpha (1 - \rho)}{2 }  ||w||^2_2,
$
where $w$ is the vector of weights $w = (w_1, ..., w_m)$, and $\alpha$ and $\rho$ are the parameters that control the $L_{1}$ (absolute value of coefficients magnitude) or $L_{2}$ (squared value of coefficients magnitude) penalisation. We chose ElasticNet because it encourages sparsity, i.e., automatically eliminates insignificant variables, and in that way deals better with the high-dimensional satellite feature vectors \cite{maharana2018use}. We performed a grid search across the parameter space while predicting vitality, and we found these parameters to work best $\alpha=0.01$, and $\rho=0.1$.




Moreover, we also experimented with \textbf{Support Vector Regression (SVR)} \cite{chang2011libsvm} method. A grid search on the parameter space yielded RBF kernel with this set of optimal hyper parameters  $C=1.0$, $degree=3$, $\gamma=$\emph{scale}, and the constant $\epsilon=0.0001$.



Finally, we also tested an ensemble of decision trees with extreme gradient boosting \cite{friedman2001greedy}, i.e., \textbf{XGBoost regression}. On small structured datasets, such as the one we dealt with, decision tree based algorithms are considered to be among the best performing methods. Upon parameter grid search, we used $350$ weak learners, learning rate of $0.01$,  the \emph{huber} loss, and the maximum tree depth of $3$.


\subsection{Metrics}
To assess how well each of the regression methods predicted our urban variables from satellite features, we resorted to two of the standard metrics used in regression methods: coefficient of determination ($R^2$), \rebuttal{its adjusted version ($R^2_{adj}$)}, and mean absolute error (MAE). Simply, $R^2$ measures the proportion of the variance in the target variable (urban variables) that is predictable from the input variables (image features), \rebuttal{while its adjusted version $R^2_{adj}$ accounts for the number of input variables and training data size, hence preventing a spurious increase in $R^2$ that is only due to the introduction of new input variables}. MAE measures the errors between true and predicted values (set of $l$ value pairs):
\begin{equation} \label{eq:mae}
MAE = \frac{1}{l} \sum_{j=1}^l | y_j - \hat{y_j} |.
\end{equation}


\begin{figure*}
\centering

	\begin{subfigure}[t]{0.91\textwidth}
	\centering
	\includegraphics[width=.82\linewidth]{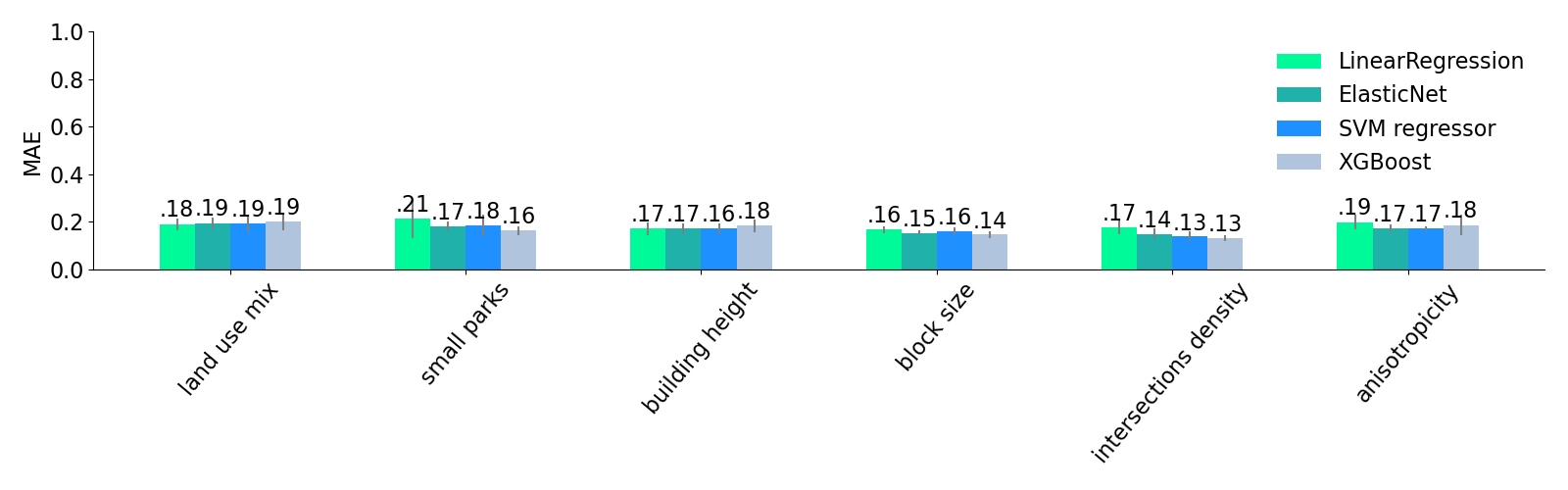}
	\includegraphics[width=.82\linewidth]{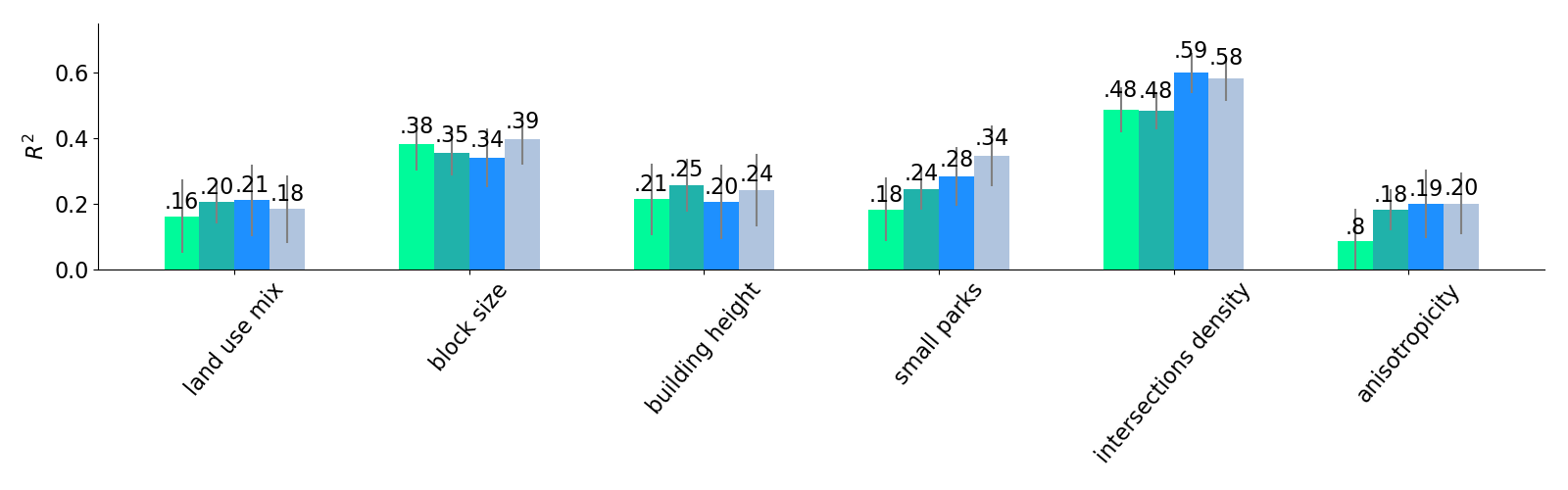}
	\caption{CAE features}
\end{subfigure}%
\hfill
\begin{subfigure}[t]{0.91\textwidth}
	\centering
	\includegraphics[width=.82\linewidth]{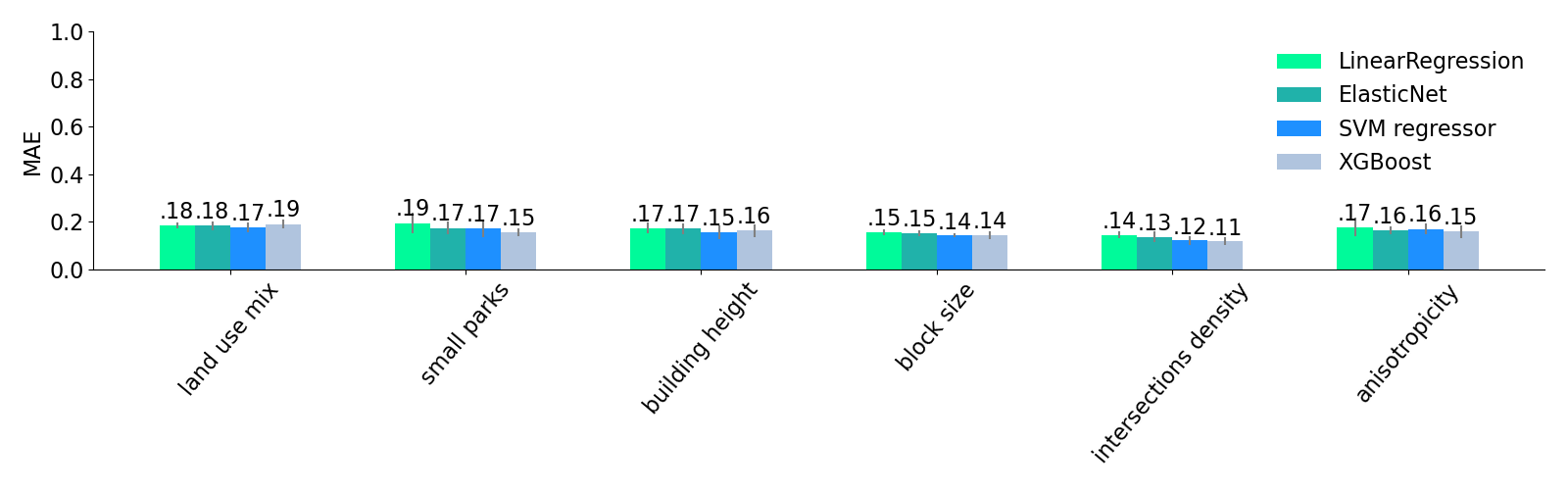}
	\includegraphics[width=.82\linewidth]{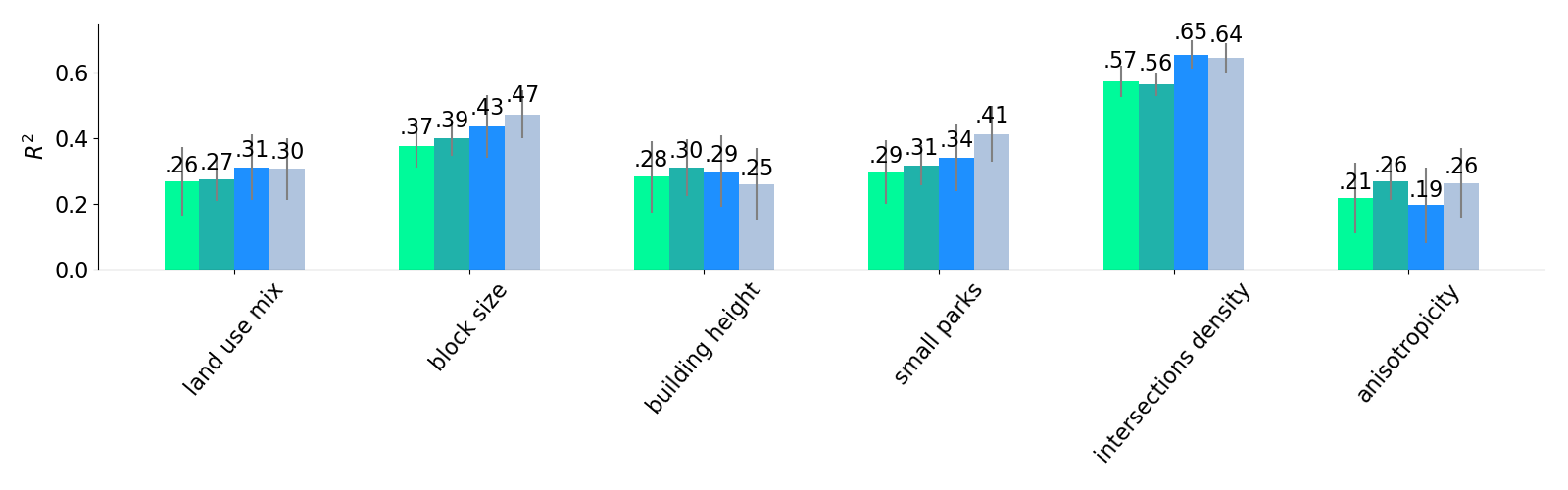}
	\caption{$VGG16$ features}
\end{subfigure}%
\caption{Regression models predicting the six vitality proxies in Table~\ref{table:selected_JJ_vars}  from satellite-derived features: (a) CAE-based, (b) $VGG16$-based features. This is the first stage of our two-step regression model, where we predict the features first, before predicting vitality. Average results of 5-fold cross-validation ran 100 times are shown: (Top) Mean absolute error (MAE) scores; (Bottom) Coefficients of determination ($R^2$). Error bars represent standard deviation across the runs. \label{fig:res_JJ_var}}
\end{figure*}

\subsection{Predicting the Six Vitality Proxies}\label{sec:predict_JJ_vars} 
We applied the selected regression methods to estimate each of the six proxies in Table \ref{table:selected_JJ_vars},  and then used the estimated values to predict urban vitality with a two-stage regression. For each of the methods, we conducted a 5-fold cross-validation 100 times (repeated k-folds) and averaged the scores. This is done to obtain stable results given a high variance from our low sample size.

The results of the first stage are shown in Figure~\ref{fig:res_JJ_var}. The starting observation is that the different features ($VGG16$ vs CAE-based) yield comparatively similar patterns across the variables, i.e., they both predict \emph{intersection density} and \emph{block size} best. However, the $VGG16$ features performed better overall for all the variables, despite the fact that the CAE was trained on satellite imagery and $VGG16$ on regular imagery. It is likely that this is due a relatively small size of our dataset (9K imagelets, in total) and that CAE features can become competitive if a larger dataset for training would be used. Hence, from now on, we discuss the results for $VGG16$ features. Out of the six proxies,  \emph{anisotropicity} is the hardest to predict. SVR and XGBoost are two methods that are the best performing overall. Another variable that is challenging to predict is \emph{building height} (the $R^2$ score of SVR is $.30$ and of XGBoost $.24$). For the four other variables, SVR was able to predict them with $R^2$ score of $.31$ for \emph{land use mix}, $.43$ for \emph{block size}, $.33$ for \emph{small parks}, and up to $.65$ for \emph{intersection density}.  They are easier to predict  not least because they can be seen in the images by the naked eye.


In the second stage, we studied whether we can predict vitality indirectly from the previously estimated six proxies. In particular, we considered all the selected variables, took their predicted vectors and applied the same regression models to predict vitality.  We found (results in Table \ref{table:indirect_all}) that the best performing method in this setup was the form of linear regression, i.e.,  ElasticNet ($R^2 = .37 \pm .07$).  



\begin{table}[t!]
	\caption{Regression results for predicting vitality from the previously inferred six vitality proxies. 5-fold cross-validation results are shown. This is the second stage of our two-step regression model.  \label{table:indirect_all}}
	\footnotesize
	\begin{tabular}{lll}
		\hline
		\textbf{Method} &  \textbf{R2}    \\
		LinearRegression &     $.336 \pm .063$ \\ 
		ElasticNet &  $ .366 \pm   .070$ \\
		SVM regressor  &   $ .357   \pm       .064$ \\
		XGBoost & $.228 \pm  .097$ \\
		\hline
	\end{tabular}
\end{table}

Finally, we studied the relative importance of the six variables in predicting vitality. By looking at the coefficients found by ElasticNet for each of those variables (Table \ref{table:indirect_vitality}), we found that the most predictive variable positively influencing vitality is intersection density (coefficient $.497$), which is in line with the previously reported results by \citet{de2016death}. Also \emph{small parks} is positively influencing vitality, while \emph{block sizes} negatively influences vitality. The \emph{building height}, which is defined through the average number of floors in buildings, matters as well: the higher it is, the less chances for creating services, such as restaurants and stores at the ground floor -- and, hence, the lower the vitality.

\begin{table}[t!]
	\caption{ElasticNet coefficients for predicting vitality from the six proxies estimated from satellite imagery.\label{table:indirect_vitality}}
	\footnotesize
	\begin{tabular}{ll}
		\hline
		\textbf{Variable} &  \textbf{coef}   \\
		intersections  density &     0.497 \\
		small parks &   0.124 \\
		anisotropicity &              0.063 \\
		land use mix  &    0.019 \\
		block size  &    -0.118 \\
		building height     &     -0.152 \\
		\hline
	\end{tabular}
\end{table}


\begin{figure*}
\centering
\includegraphics[width=.65\linewidth]{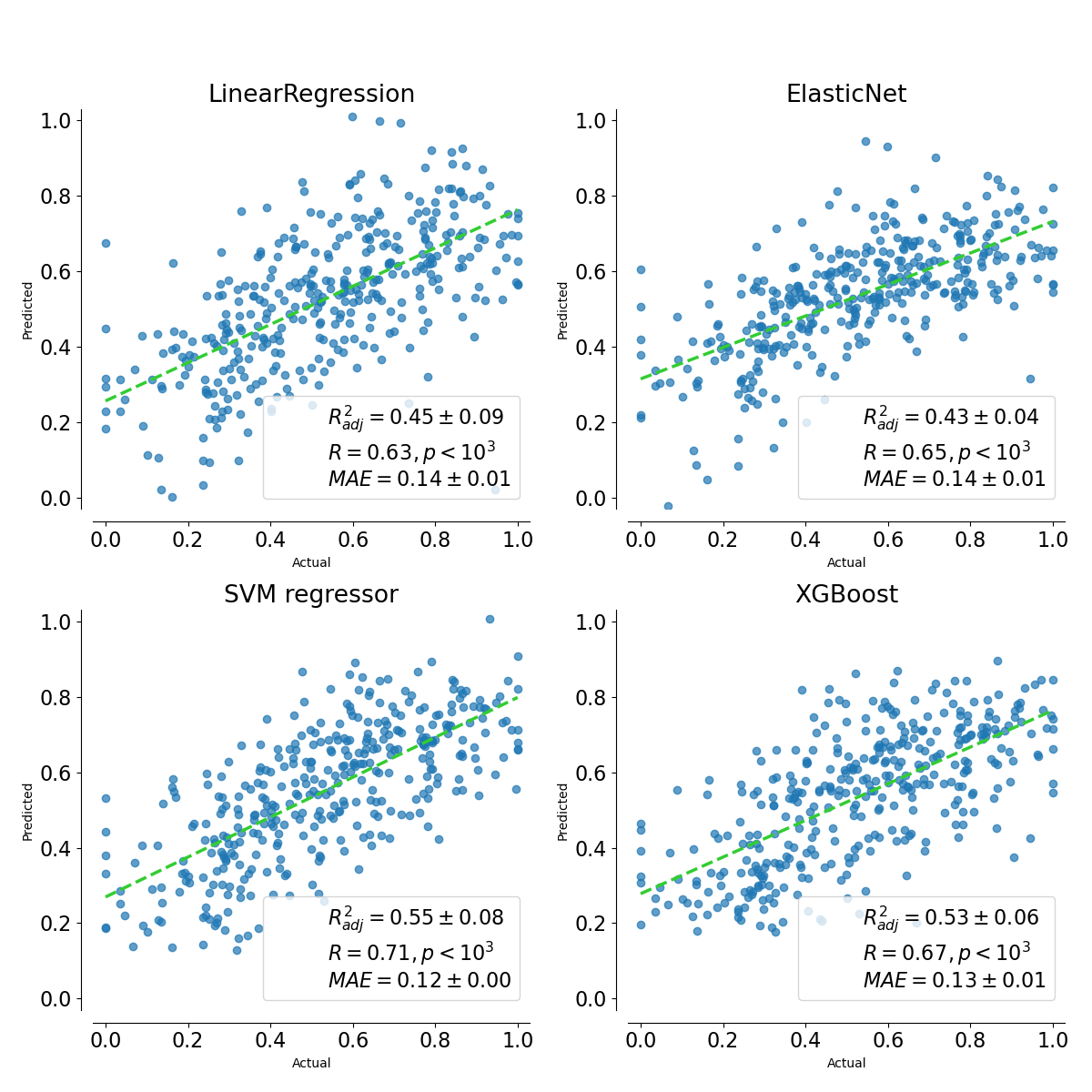}
\caption{Regression models that predict vitality: 5-fold cross-validation results.  Each plot represents a different model \rebuttal{evaluated in terms of: adjusted coefficient of determination ($R_{adj}^2$), Pearson correlation coefficient ($R$) ($p<.001$), and mean absolute error ($MAE$)}. To aid visual interpretation, \rebuttal{a green line representing the best linear fit between true and predicted values is shown.}\label{fig:res_vitality}}
\end{figure*}

\begin{figure}
\centering
\includegraphics[width=.82\linewidth]{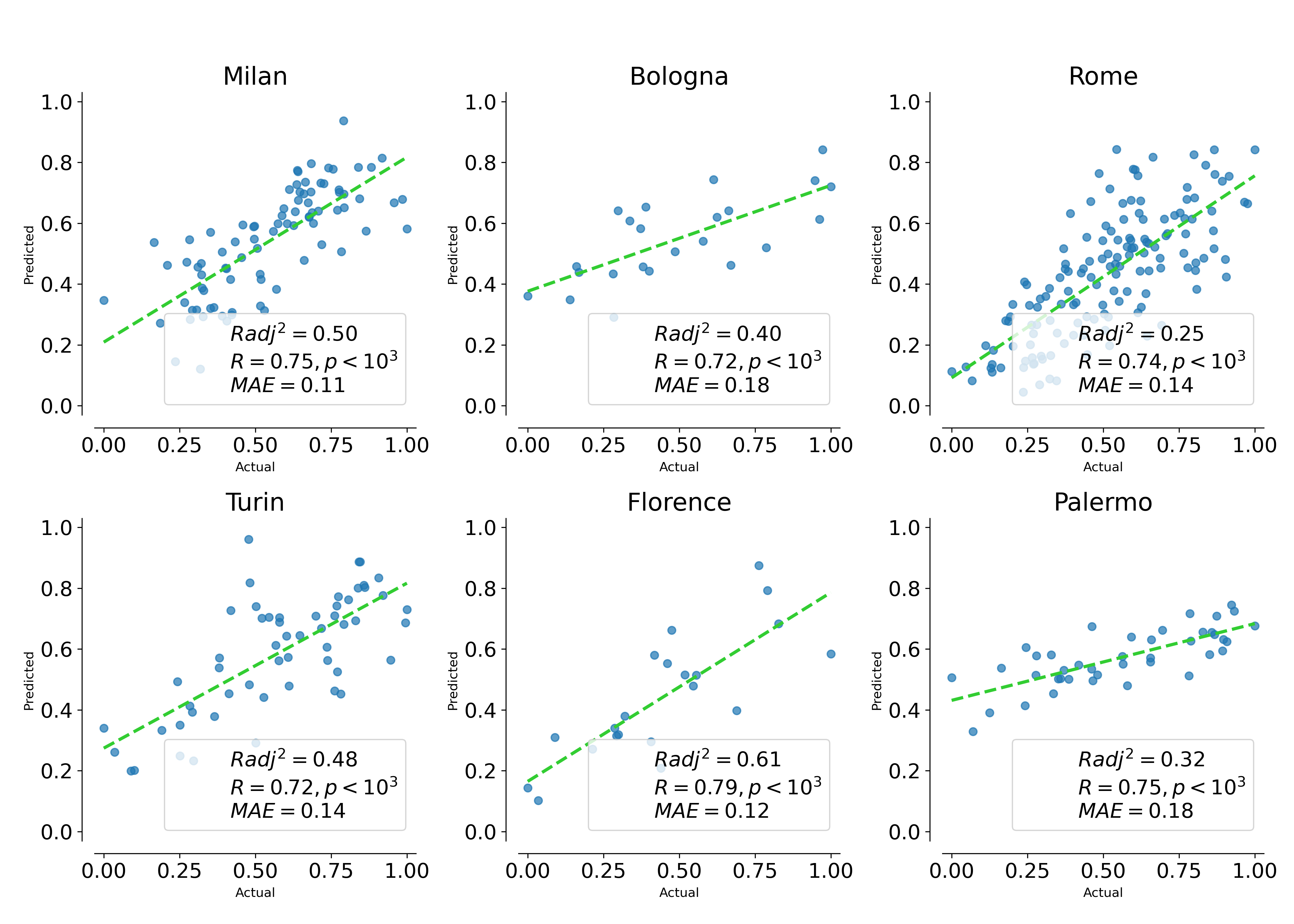}
\caption{Leave-one-city-out prediction results using SVR. The model is trained on the data from 5 cities and tested on the last city.  Each plot represents the model for a different city (the one on which it was tested) \rebuttal{evaluated in terms of: adjusted coefficient of determination ($R_{adj}^2$), Pearson correlation coefficient ($R$) ($p<.0001$), and mean absolute error ($MAE$)}. To aid visual interpretation, \rebuttal{a green line representing the best linear fit between true and predicted values is shown.} \label{fig:different_cities}}
\end{figure}

\subsection{{Predicting Vitality}}\label{sec:predict_vitality}
In our second experiment, we evaluated the predictive power of the four regression models in estimating vitality directly from the visual features of the satellite imagery. First, for each of the models, we conducted a 5-fold cross-validation experiment (Figure \ref{fig:res_vitality}). In the previous set of the experiments, SVR and XGBoost regressor were the best models and shown a similar performance. In this experiment, SVR ($R^2_{adj} = .55 \pm .03$, $MAE =  .12 \pm .00$) slightly outperformed XGBoost ($R^2_{adj} = .54 \pm .04$, $MAE =  .13 \pm .01$). 

\begin{figure*}
	\centering
	\begin{subfigure}[t]{0.33\textwidth}
		\centering
		\includegraphics[width=.99\linewidth]{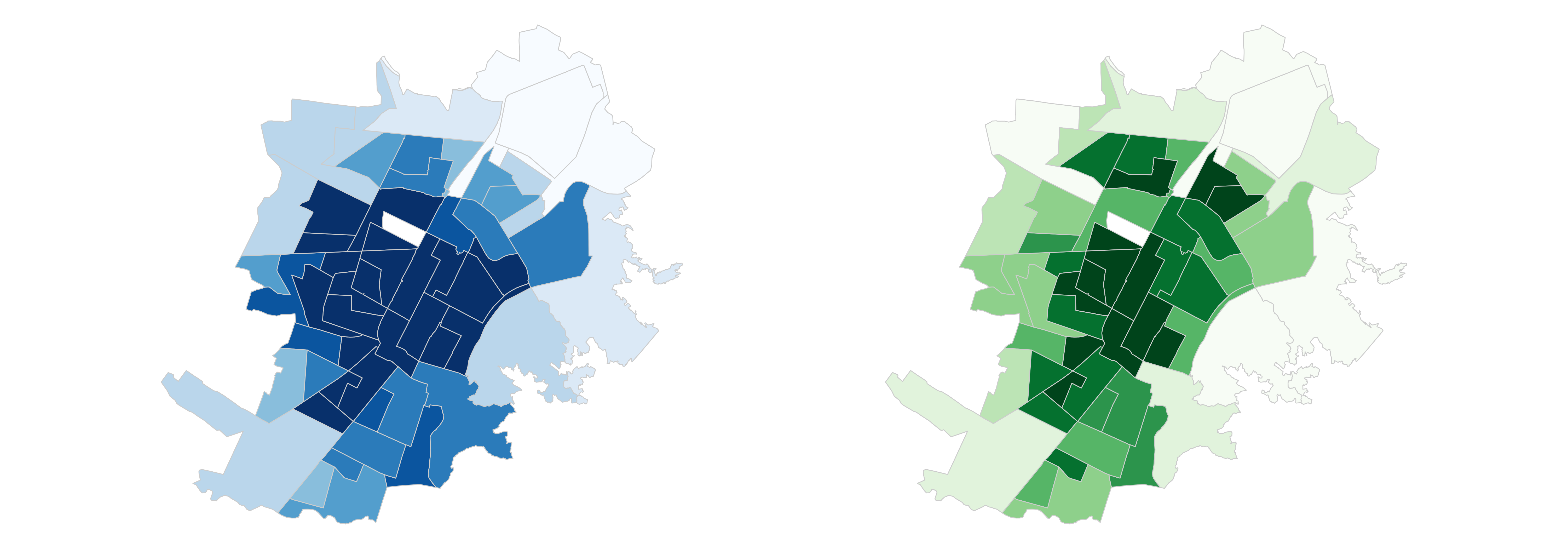}
		\caption{Turin}
	\end{subfigure}%
\begin{subfigure}[t]{0.33\textwidth}
	\centering
	\includegraphics[width=.99\linewidth]{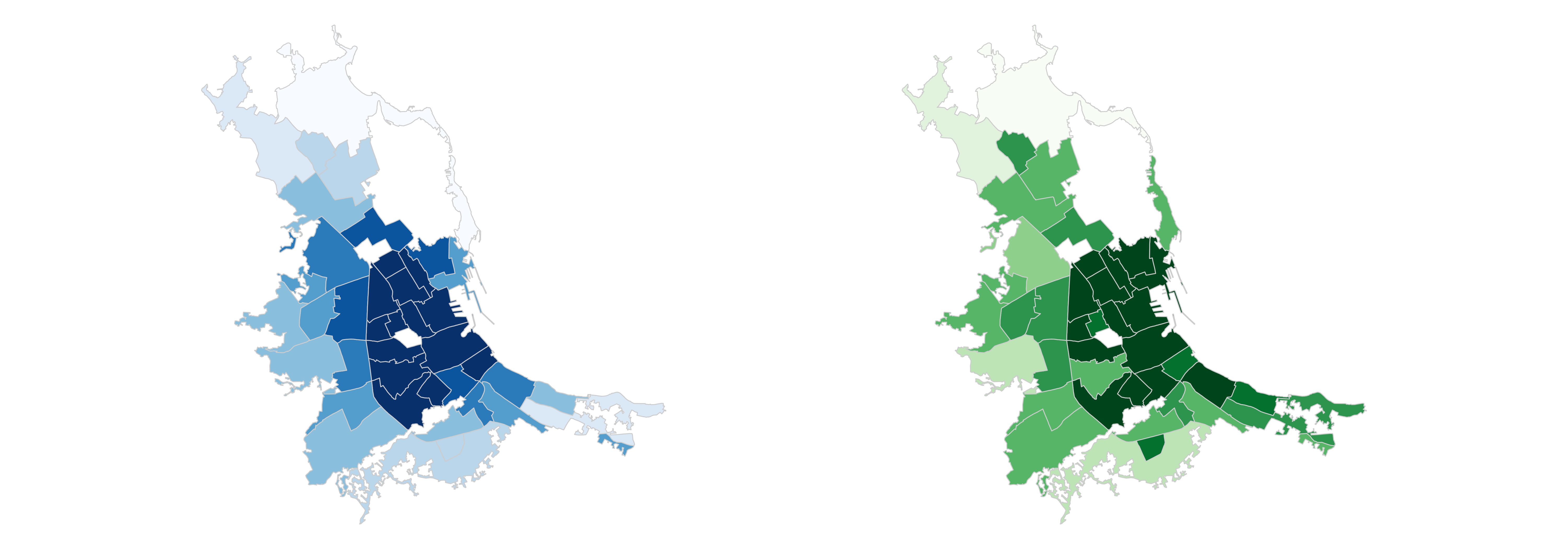}
	\caption{Palermo}
\end{subfigure}%
\begin{subfigure}[t]{0.33\textwidth}
	\centering
	\includegraphics[width=.99\linewidth]{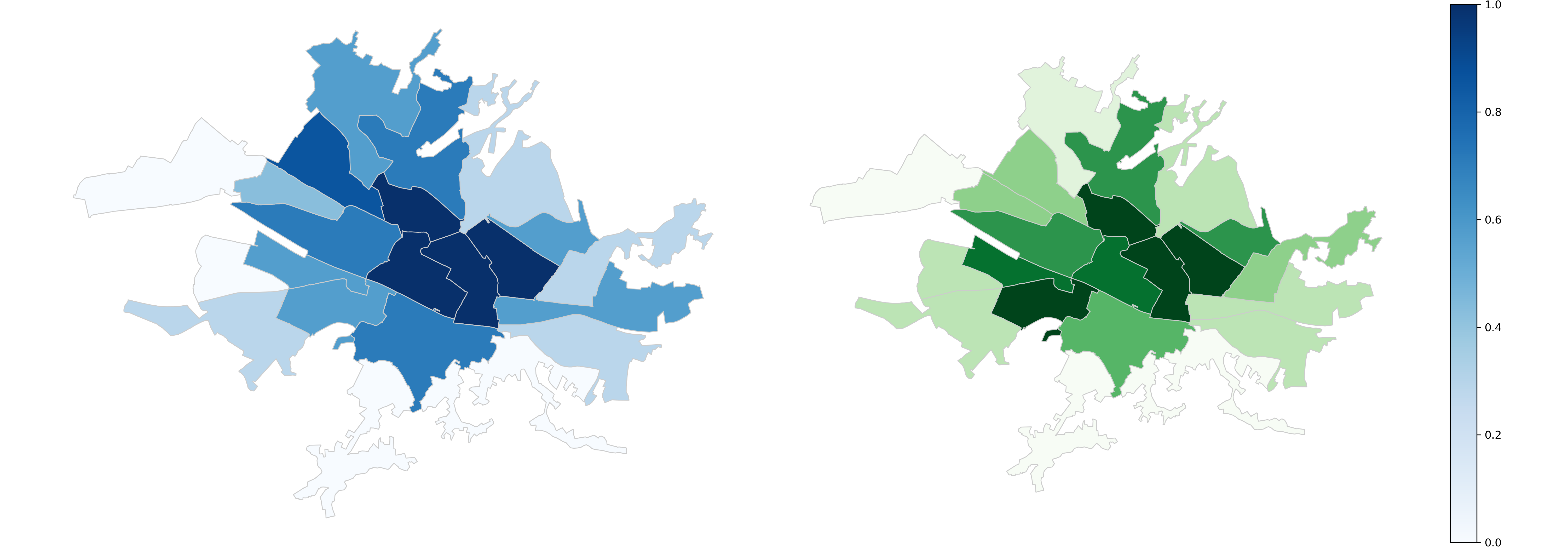}
	\caption{Florence}
\end{subfigure}

\begin{subfigure}[t]{0.33\textwidth}
	\centering
	\includegraphics[width=.99\linewidth]{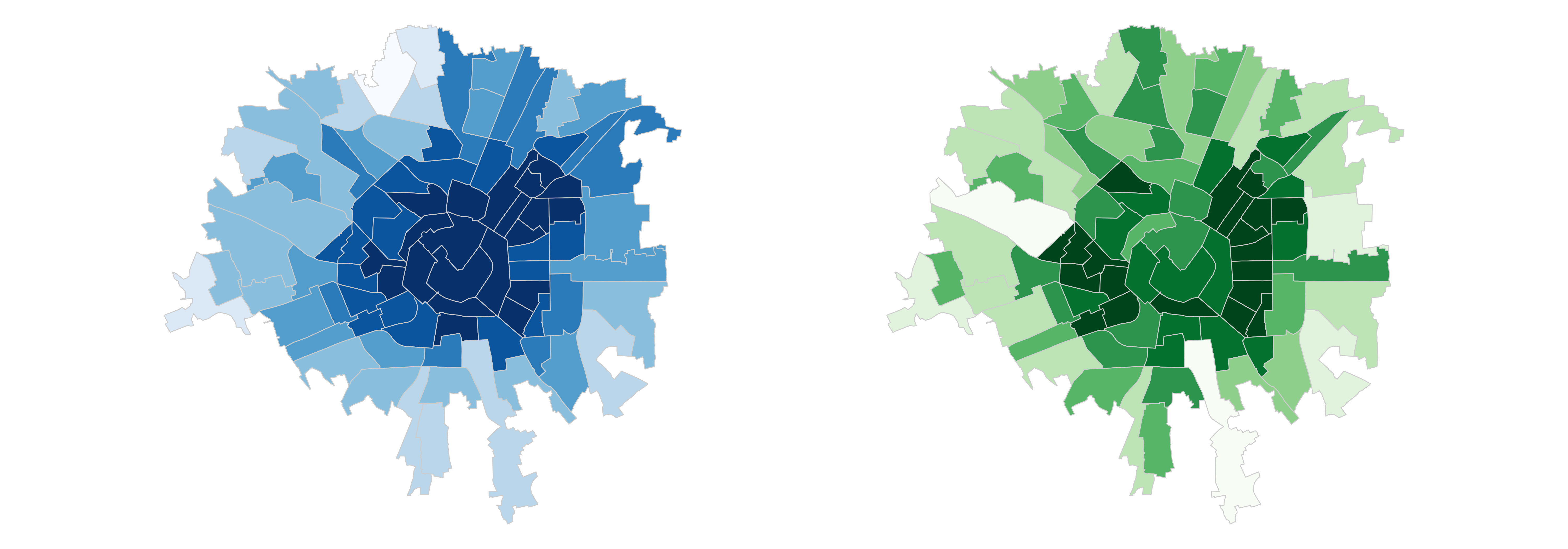}
	\caption{Milan}
\end{subfigure}%
\begin{subfigure}[t]{0.33\textwidth}
	\centering
	\includegraphics[width=.99\linewidth]{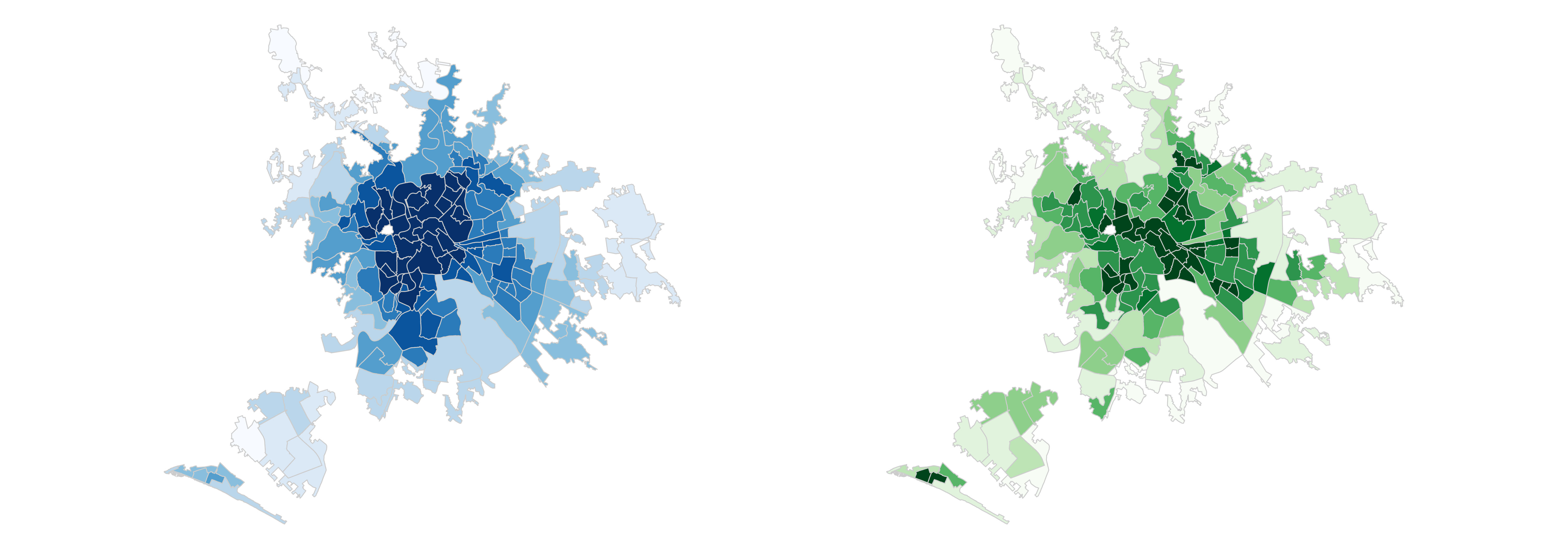}
	\caption{Rome}
\end{subfigure}%
\begin{subfigure}[t]{0.33\textwidth}
	\centering
	\includegraphics[width=.99\linewidth]{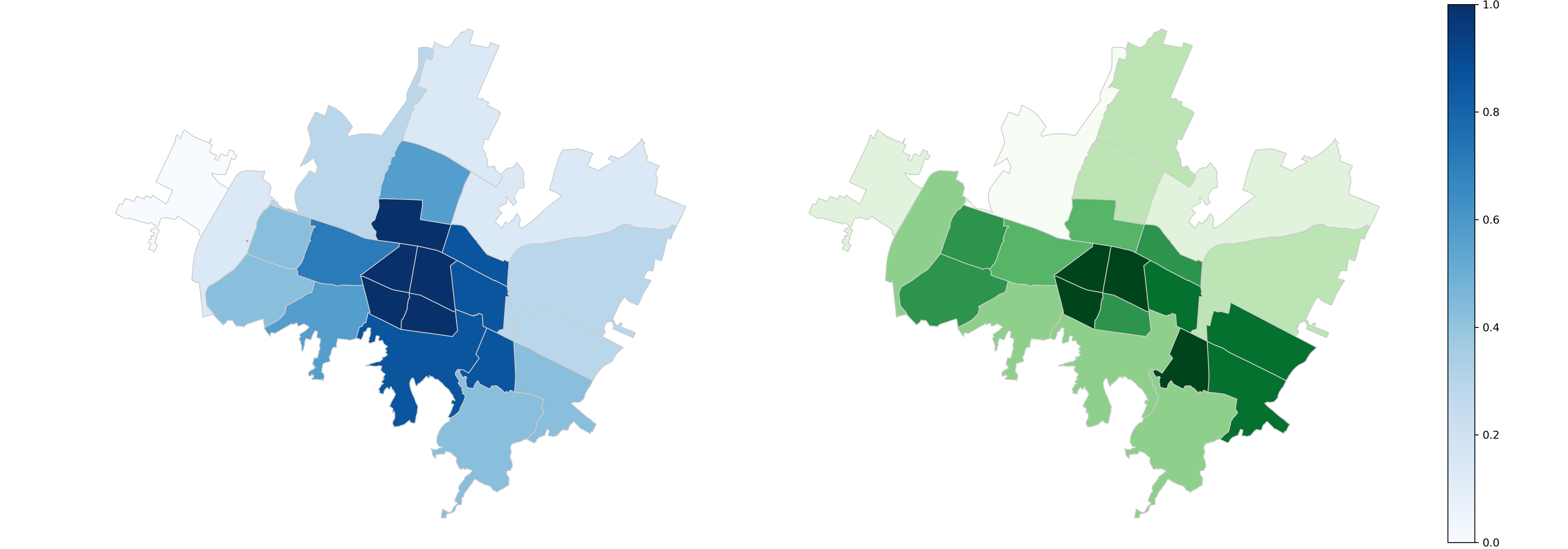}
	\caption{Bologna}
\end{subfigure}%
	\caption{\rebuttal{Maps of the true (blue) and predicted (green) urban vitality levels. Across the cities, the predictions follow  an overall pattern of higher vitality in the centers, and lower vitality in the suburbs. However, inconsistencies exist for specific cities. Notably, in Milan, predictions for the districts in the very center of the city (containing two large parks) are high but not maximum. On the other hand, our model overestimates vitality for Bologna's East districts, which, again, happen to feature small parks and relatively small blocks.  \label{fig:maps}}}
\end{figure*}

In the final experiment, we studied the generalizability of the urban vitality prediction, i.e., we asked whether the models can be trained on some cities to predict vitality in another (unseen) city. For this purpose, we conducted leave-one-city-out validation experiments and used SVR, as it was the best performing model in the previous experiments. Specifically, for each of the six cities, we trained the model on the data from five cities and tested it on the remaining city. The results are shown in Figure \ref{fig:different_cities}. 
The model's predictive power varies: in Milan and Florence, it is able to explain, respectively, \rebuttal{$50\%$ and $61\%$ of the variance in vitality, but, in Palermo and Rome, the explained variance is, respectively, $32\%$, and $25\%$.} 
To illustrate the ability of satellite-derived features in predicting vitality, in Figure \ref{fig:maps}, we show maps with true and predicted vitality for \rebuttal{the six cities.}


\subsection{\rebuttal{Explaining Vitality and its Proxies}}
\label{sec:explaining}
\rebuttal{Our analyses up to this point suggested that satellite data can predict urban vitality, and it can do so with varying degrees of accuracy. Our best performing method (SVR), tested across all the cities, achieved an $R_{adj}^2$ of $.55$, and its scores on individual cities ranged from $.61$ for Florence to $.25$ for Rome.}
\rebuttal{
	In this section, we study the factors potentially affecting prediction accuracy (Subsection \ref{sec:factors}), and the association between models' inference and the presence of PoIs in the area (Subsection \ref{sec:POIS}).}

\begin{figure}
	\centering
	\includegraphics[width=.77\linewidth]{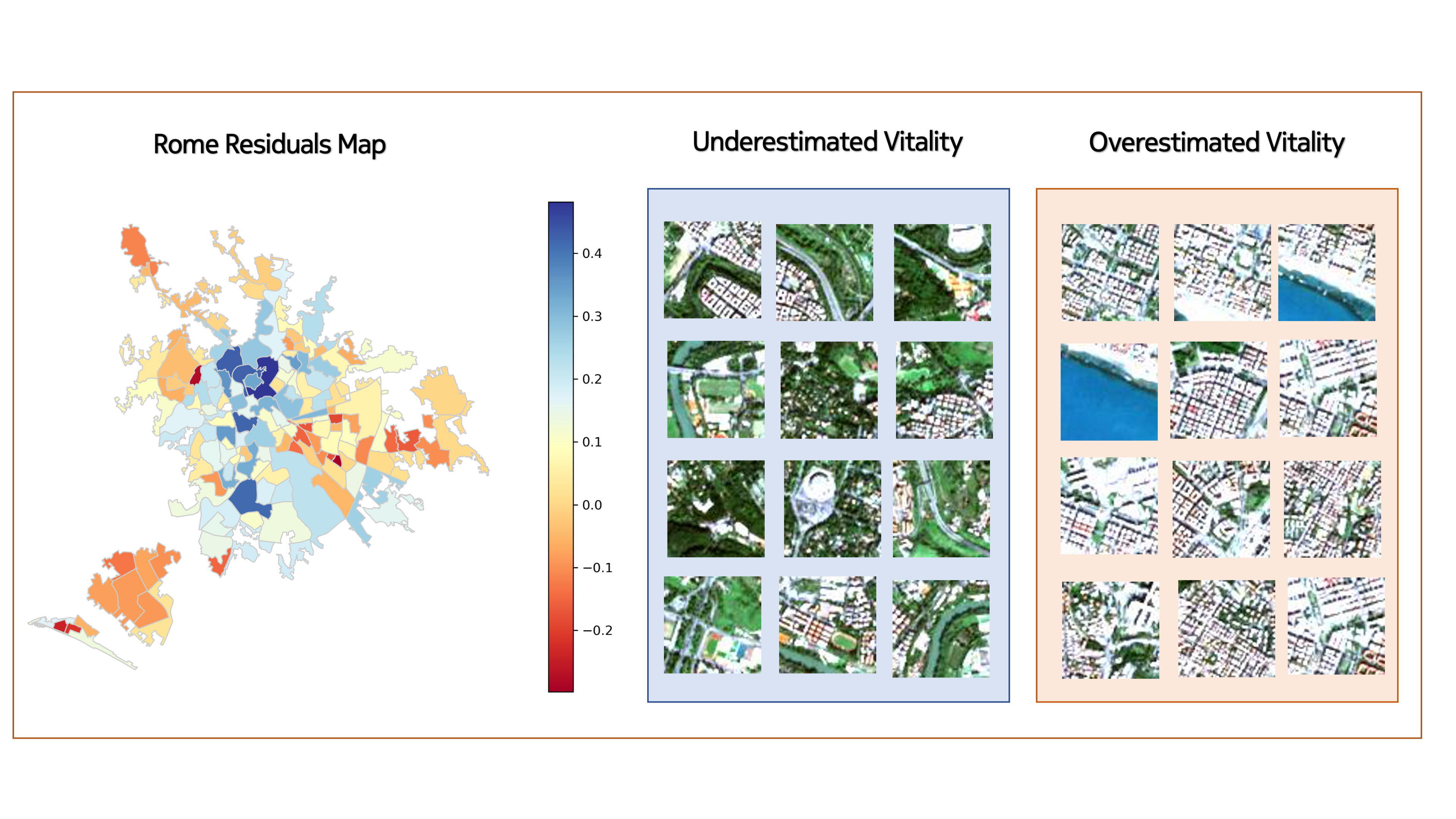}
	\caption{\rebuttal{The map of Rome's residuals and its satellite views for which our model overestimated and underestimated vitality level. The map shows residuals, i.e., differences between the true and predicted vitality levels: blue/red means that our model estimated lower/higher vitality levels compared to the ground truth. Example imagelets reveal that our model underestimated vitality levels in areas with large parks, rivers, highways, and stadiums, while it underestimated vitality levels in areas with high density of buildings and near the sea. \rebuttal{\label{fig:rome_res}}}} 
\end{figure}

\subsubsection{Factors affecting vitality prediction}
\label{sec:factors}
\rebuttal{The city with the lowest prediction score in our experiments is Rome ($R_{adj}^2=.25$). That could be partly because it has the smallest training set. To see why, consider that Rome has the \emph{largest number of districts} (see Table \ref{table:stats_cities}) and, in our  leave-one-city-out evaluation, Rome's training set consists of areas in the remaining cities, while the test set consists of areas in Rome. As a result, Rome has the  \emph{smallest training set} compared to the other cities. So one might speculate that the limited training data could influence the adjusted $R^2$ score, i.e., $R^2_{adj}$. Indeed, Rome's unadjusted $R^2$ is $.39$, which is significantly higher than its $R^2_{adj}$.}
 \rebuttal{To study whether this is the case for the other cities, we set out to determine whether prediction performance ($R_{adj}^2$) depends on the size of the training data. To obtain a city's training data size (denoted as $TS$), we summed the number of districts $N_c$  from the remaining five cities (see Table \ref{table:stats_cities}). There was no significant correlation between $TS$ and $R_{adj}^2$ across cities, meaning that the prediction score was generally not influenced by the size of the training data.}
	

\paragraph{Prediction residuals for Rome.} To look for alternative explanations, we further focused on Rome. Figure \ref{fig:rome_res} shows the map of its residuals, i.e., the differences between true and predicted vitality levels. The areas in blue/red are those for which our model underestimated/overestimated. \rebuttal{We found that our model tended to \emph{underestimate} in: \emph{i)} areas with large parks and green spaces, rivers, highways, and stadiums (i.e., urban features that Jane Jacobs named \emph{border vacuums}); and \emph{ii)} areas with ancient landmarks in the city center. Jane defined border vacuums as \emph{``perimeters of a single massive or stretched-out use of territory [...] that exerts an active influence on pedestrian activity''}. According to her theory, border vacuums can, indeed, depending on the context, be associated with both high vitality (\emph{``if some visual or motion penetration is allowed through it, [...] it then becomes a seam rather than a barrier, a line of exchange along which two areas are sewn together''}) and low vitality (if the border vacuum cuts off activity in public spaces). Our analyses reveal that border vacuums are  associated with high vitality in central parts of Rome, and with low vitality in the other cities  (for example, Figure \ref{fig:TorinoLabels} shows border vacuums in low-vitality areas of Turin). Also, the overall percentage of greenery in Rome is higher than that of any of  the other cities\footnote{The percentage of greenery is $40\%$ in Rome, compared to less than $13\%$ in Milan. \url{http://www.worldcitiescultureforum.com/data/of-public-green-space-parks-and-gardens}}.}
\rebuttal{By inspecting the underestimated central locations on Google Maps, we also found that they included ancient landmarks popular among tourists (e.g., Largo di Torre Argentina, Foro Romano). These  landmarks are not urban features typically found in the other cities (especially not in Turin where the size of its Roman district is quite limited), and, based on satellite images, might appear to be similar to ruins. That might be why the model trained on the cities other than Rome did not learn to label those locations in Rome as being of high vitality, despite being popular among tourists.}
	

On the other hand, we found that our model  tended to \emph{overestimate} vitality levels in areas with small blocks and dense housing predominantly located in  the south-west district near the sea (the City of Fiumicino), which  hosts the busiest airport in Italy. As one expects, given noise pollution, this district does not enjoy a rich outdoor  life (including the presence of retail shops) as more central districts do, impacting one of the four Jacobs' dimensions, that of mixed economic activities. 

\crc{
\paragraph{Prediction residuals for Turin.} To extend our prediction factors analysis to a different geographical, cultural, and economic environment compared to Rome, and also to look at a city where our models performed better, next we inspected Turin. Turin is located at the north of the country, does not have an access to the seaside like Rome; it has just one third of Rome's population, but due to the smaller size, it is three times more densely populated. Turin features a beautiful historical city centre, and buildings, castles, public squares in architectural styles ranging from renaissance, to rococo, and art nouveau. The city also hosts headquarters of several well-known Italian car-maker companies, and two large football stadiums.}

\begin{figure}
	\centering
	\includegraphics[width=.77\linewidth]{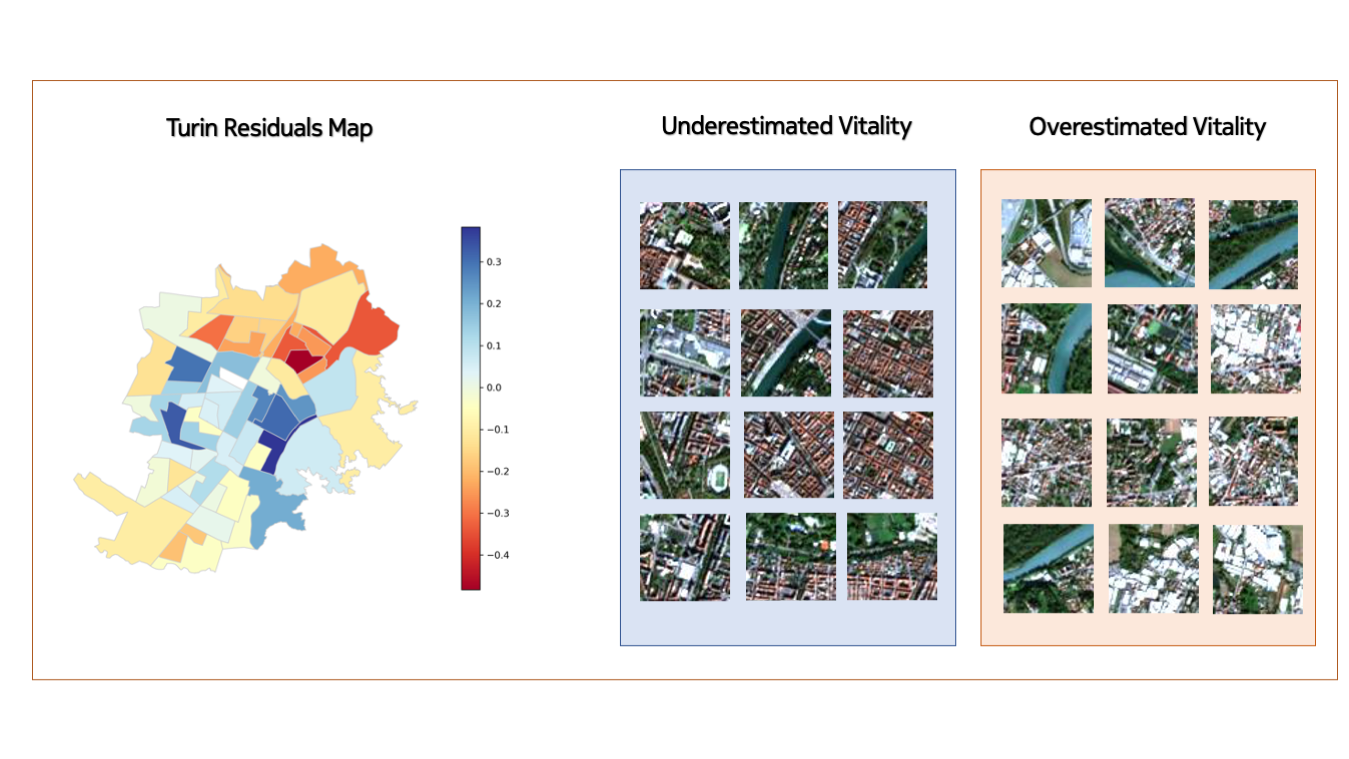}
	\caption{\rebuttal{\crc{The map of Turin's residuals and its satellite views for which our model overestimated and underestimated vitality level. The map shows residuals, i.e., differences between the true and predicted vitality levels: blue/red means that our model estimated lower/higher vitality levels compared to the ground truth. Example imagelets reveal that our model overestimated vitality levels in areas with large industrial buildings and close to the natural park, while it underestimated vitality levels in areas among the three popular Turin city parts (Porta Nuova, Parco del Valentino, and Giardini Reali Superiors). \label{fig:turin_res}}}}
\end{figure}

\crc{
	Figure \ref{fig:turin_res} shows the map of residuals for predicted vitality levels in Turin. This time, too, the areas in blue/red are those for which our model underestimated/overestimated vitality. We found that in Turin, our model tended to \emph{underestimate} in: \emph{i)} areas in between the largest Railway Station in the city, Porta Nuova, to the west, and a popular public park, Parco del Valentino, to the east, and up to the elegant park with statues and kid's play equipment, Giardini Reali Superiors, to the north, and \emph{ii)} in areas next to the smaller athletic stadium, Stadio Primo Nebiolo. In the area among the three popular Turin city parts (Porta Nuova, Parco del Valentino, and Giardini Reali Superiors) that are at a walking distance from each other, indeed, the pedestrian mobility is maximally high. However, from the satellite imagery showing a mix of buildings, rivers, and relatively-large parks, our models inferred it to be high but still not the maximum level. The area around the athletic stadium Stadio Primo Nebiolo, interestingly, is associated with high vitality, while, for instance, the Juventus Stadium is found in a low-vitality area to the north-west. This, once again, confirms that border vacuums can have different interplay with vitality, and, in this specific case, that has caused our method to underestimate vitality around Stadio Primo Nebiolo.
}

\crc{On the other hand, we found that our model tended to \emph{overestimate} vitality levels in the northeast areas of Turin, close to the river Po and the nature park, Riserva Naturale del Meisino e dell'Isolone Bertolla. To the north from the park, there are several large industrial buildings, hosting agricultural production, warehouses, and some cooperatives. These buildings, as can be seen in the satellite images for overestimated vitality in Figure \ref{fig:turin_res}, appear white, and are mixed with red buildings likely representing residential houses nearby, and with parks. We speculate that our models, in this case, took this diversity as a mix of old and new buildings close to small parks, overestimating vitality for them.}

\subsubsection{Association between the model's inference and PoIs presence}\label{sec:POIS}
\rebuttal{Our prediction pipelines consist of deep learning methods that extract visual features. One issue is that the extracted features are hard to interpret and explain. 
	To understand the associations between these abstract image features and actual points of interest on the ground, we queried the OpenStreetMap (OSM) database and collected all the Points of Interest (PoI) across the 6 Italian cities. OSM classifies the PoIs into 6 broad categories of amenities\footnote{\label{note1}https://wiki.openstreetmap.org/wiki/Key:amenity}: \textit{sustenance, education, transportation, financial, healthcare,} and \emph{entertainment}. Since vitality is a measure of human traffic in an area, we focused on the categories of sustenance, transportation, and entertainment, which cover the majority of amenities contributing to a city's social life. Next, we mapped the locations of the PoIs onto the $2{,}146$ imagelets described in Section \ref{sec:combined}. With this mapping, we were able to count in each imagelet the total number of PoIs belonging to either of the three PoI categories. 
	For any given imagelet $i$ containing a total number of  $n_{sustenance}$, $n_{transportation}$, and $n_{entertainment}$ PoIs, 
	we computed the PoI scores for each category:
	\begin{equation}
	\label{eq:catScores}
	S^{i}_{cat} = \log(1+n_{cat}), \quad \forall {cat} \in \{sustenance, transportation, entertainment\}
	\end{equation}
	The log operation allowed us to transform the long tailed distribution of the number of PoIs into a normal distribution. We added 1 to avoid undefined scores for imagelets with zero PoIs of any given category. We then used these three scores computed for all the imagelets to explain the association between the model's inference and the PoI categories on the ground.}

\begin{figure}
	\centering
	\includegraphics[width=.7\linewidth]{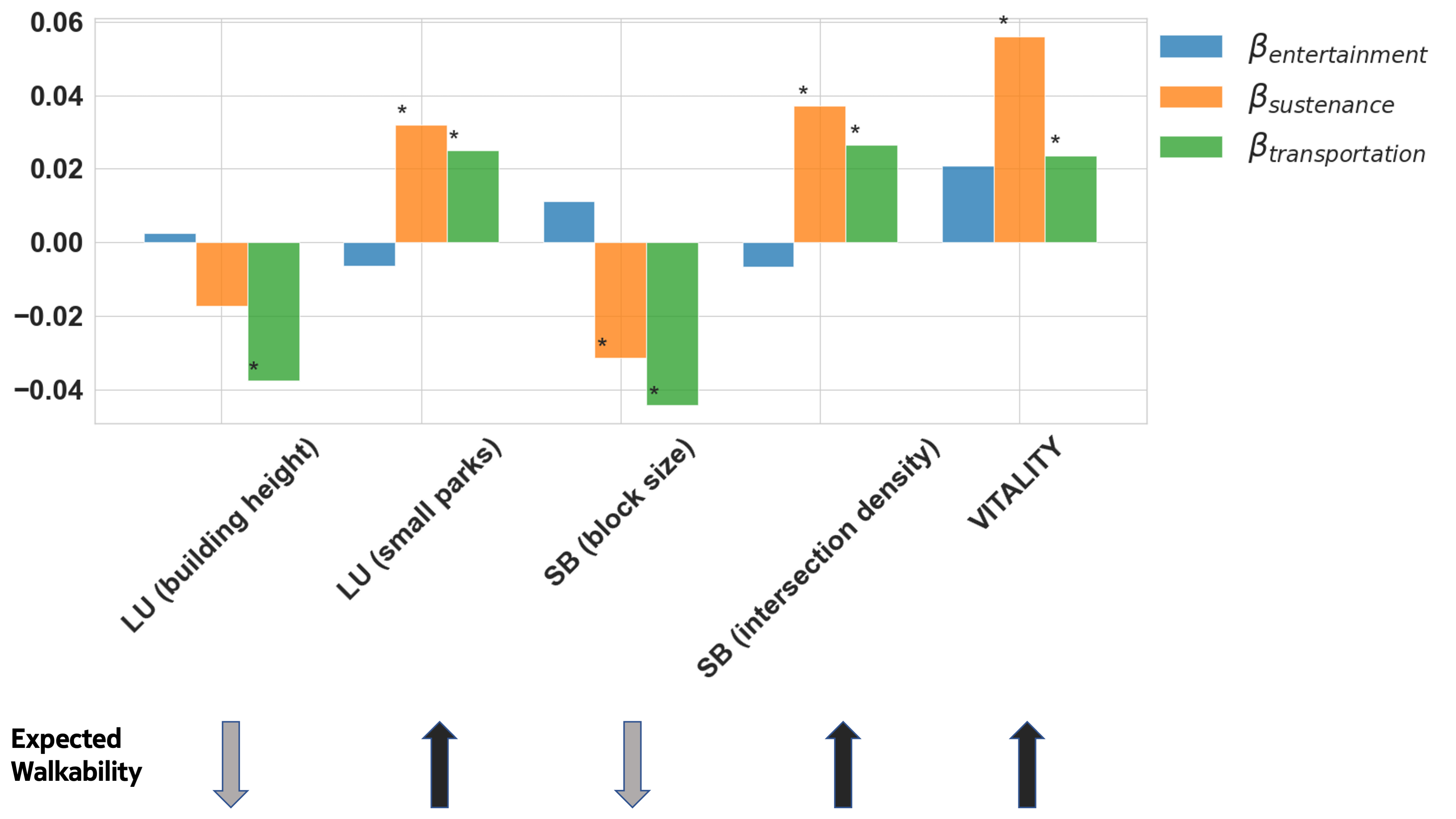}
	\caption{\rebuttal{The logistic regression (LR) coefficients for each OSM Places category in imagelets with high/low values (as output by our models based on $VGG16$-based features) for the six urban variables. The included variables are the five vitality proxies for which prediction on our satellite data with both types of features yielded $R^2$ above $.25$ (see Figure \ref{fig:res_JJ_var}) and vitality. Significant LR coefficients ($p$-value $< .05$) are marked with a $*$. The up/down arrows denote the increasing/decreasing expected association between a specific urban variable (e.g., building height), and walkability/pedestrian presence in an area. \label{fig:POIs}}}
\end{figure}

\rebuttal{We did this by splitting the imagelets into two groups, a separate split each time, for vitality and each of its proxies, based on their ground truth values. 
The first group contained all those imagelets whose ground truth \crc{values} for vitality, or one if its proxies, were in the top tertile, which we called the \textit{high} class. Similarly, the other group contained imagelets with values in the bottom tertile, which we called the \textit{low} class. We then sampled 50\% of the \textit{high} and \textit{low} groups into a training set of imagelets and kept the remaining ones as the test set. We then trained a set of binary classifiers to predict vitality and its proxies using the satellite-derived features on the training set. The models' performances were comparable to the original models', with AUC scores in a 5-fold cross validation setup equal to $.63 \pm .07$ for small parks, $.88 \pm .06$ for small blocks, and $.93 \pm .06$ for vitality. 
We then predicted the binary classes for vitality and its proxies for the test set of imagelets using these trained models. 
Finally, to quantify the relationship between the predicted labels on the imagelets and the PoIs found on the ground, we binarized the predicted class labels, i.e., class $c_i = 1$, if imagelet $i$ is predicted to be in the \textit{high} class, and $c_i = 0$, if imagelet $i$ is predicted to be in the \textit{low} class. 
We also computed the category scores for any given imagelet $i$ using equation (\ref{eq:catScores}). These scores are denoted by $S^i_{sustenance}$ , $S^i_{transportation}$ and $S^i_{entertainment}$, and were normalized on the $[0,1]$ scale.
With this setup, we then fit a logistic regression that predicts the probability of an imagelet being in class 1, given the OSM PoI category scores.} 
\begin{equation}
Pr(c_i =1) = \frac{1}{1+ e^{-(\alpha + \beta_1 \cdot S^i_{entertainment} + \beta_2 \cdot S^i_{sustenance} + \beta_3 \cdot S^i_{transportation} )}}\\
\label{eq:logit}
\end{equation}
\rebuttal{Figure \ref{fig:POIs} shows the values for the logistic regression coefficients $\beta_1$, $\beta_2$, $\beta_3$ associated with the predictors $S_{sustenance}$ , $S_{transportation}$, and $S_{entertainment}$, respectively. One can interpret the values of the $\beta$ coefficients in terms of odds ratios.  To get an upper bound of the predictive difference corresponding to a unit difference in each predictor, we used the ‘divide by 4’ rule~\cite{gelman2007analytical} (p. 82). This says that, in a logistic regression, by dividing each predictor’s coefficient by 4, one can draw conclusions about the increase in likelihood $Pr(c_i = 1)$ for each unit increase in the predictor. Simply put, we can quantify the percentage increase in the probability of the dependent variable to be in the positive class.} \rebuttal{Indeed, the results in Figure \ref{fig:POIs} reveal that satellite features capture meaningful insights in terms of places: all three place categories are positively predictive of vitality. Specifically, each 1\% increase in the PoI score for the sustenance category increases the likelihood of the imagelet to belong in the \textit{high vitality class} (+1.5\%) and the \textit{high intersection density} class ($\approx +1$\%). Similarly, a 1\% increase in the PoIs score for the transportation category would decrease the corresponding imagelet's probability of belonging to the \textit{high block size} class (-1\%). PoIs in the sustenance and transportation categories are both predictive of \emph{higher intersection density} and more likely presence of  \emph{small parks}.  Also, as expected, these two categories are negatively associated with \textit{high block size} (large blocks, which do not encourage walkability), and \textit{high building height} (high-rise buildings). {In summary, the results associated with PoIs from the sustenance and transportation categories meet expectations. For the entertainment category, however, this is only the case for  vitality and for its proxy ``block size'' but not for the other proxies. That is partly because the OSM entertainment category\textsuperscript{\ref{note1}} contains places that are typically found in dense city centers characterized by high vitality (e.g., fountain, cinema, theater) and places that  act as border vacuums characterized by low vitality (e.g., casino, gambling, convention center). }}
		

\section{Discussion and Conclusion}\label{sec:discussion}

We have proposed a deep-learning framework that extracts features from publicly available Sentinel-2 imagery and predicts not only the six proxies for urban vitality (i.e., indirect prediction) but also vitality itself (i.e., direct prediction). 
\rebuttal{The framework turned out to work better for the direct prediction of vitality ($55\%$ of the variance is explained) rather than its indirect prediction via six intermediate interpretable urban proxies ($36\%$). That \crc{can be attributed to} three reasons: \emph{i)} the six proxies upon which the indirect prediction relied do not exhaustively capture vitality (e.g., mix of economic activities and concentration of people are not captured); \emph{ii)} the prediction errors suffer from the two-step process; and \emph{iii)} the direct prediction relied on raw satellite features, which may capture aspects overlooked by the indirect prediction.}

\mbox{ } \\ 
\textbf{Practical Implications.} The presented predictive ability might be beneficial for  supporting urban planning interventions in the developed world, and for sustainable development  initiatives  in the rapidly growing developing world.  Specifically, we identified four main practical implications.

\begin{description}
	\item \emph{Satellite Data in City Dashboards.} Existing city dashboards and digital services could be enhanced with satellite data  that may well fill the gap between the high frequency nature of mobile phone data and the low frequency nature of census data. We showed that some structural features of the urban environment that are important for vitality can be inferred from satellite data. Given that this data is publicly available, at a medium-level spatial resolution (10m) and continuously updated (every 5-7 days), it can be used for monitoring changes in the urban environment, to support urban planners/designers and policymakers in their decisions and planning. This is particularly relevant for developing countries, which frequently lack access to other types of urban data.
	
	\rebuttal{\item \emph{Guidelines for Urban Measurement from Satellite Data.} Our results also suggest that having highly diverse geographic data  is critical when making inference on unseen cities, especially if these new cities are sufficiently different in terms of history or culture from the cities in the training data. } \newline 
	
	\item \emph{Facilitating Rapid Urbanization.} Jacobs's emphasis on diversification is particularly relevant to today's globalization: her theories shed light on urban inequality in Africa \cite{obeng2015social}, slum clearance in preparation for mega-events \cite{greene2003staged}, as well as efficient growth, such as in Taiwan \cite{ellerman2004jane}. Our approach can provide a dynamic view of neighborhood structure and help to track issues relevant to globalization. \newline
	
	\item \emph{Digital Earth.} Extending our methods to track vitality across the globe can contribute to the interactive view of our planet that enables a shared understanding of the relationships between the physical environment and society~\cite{gore1998digital, craglia2012digital}.
\end{description}

\mbox{ } \\ 
\textbf{Theoretical Implication.} Replicating the experiments we did in Italy in other countries will require to obtain mobile phone data that can serve as a proxy for vitality in those countries. We can then learn visual features capturing vitality, which will likely uncover subtleties in how vitality is expressed across different natural and cultural environments.


\mbox{ } \\ 
\textbf{Limitations and Future Work.} 
\rebuttal{While the openness of the Sentinel-2 dataset is an advantage, its limited spatial resolution is a limitation. If higher spatial resolution data would be publicly available, our models' performance will likely improve.}

\rebuttal{We also relied on  mobile Internet data to quantify a proxy for vitality, which could be another limitation. However, since the market share of the provider (Telecom Italia) is 34\%, which is the largest in the country, we believe that such a proxy offers a close approximation of the actual footfall in the specific context of our study.}
\rebuttal{More generally though, in today's highly connected world, mobile activity is one of the best proxies for measuring presence of  people \cite{steenbruggen2013mobile,vscepanovic2015mobile,phithakkitnukoon2012socio,deville2014dynamic}, and has been successfully used to estimate urban vitality \cite{kim2018seoul,de2016death}.}

 \rebuttal{Another limitation with the data has to do with the temporal lag between mobile phone data and satellite data, which, in our case, amounts to three years. It is possible that some aspects of urban vitality could have changed during this time period.}

Moreover, since this study is only conducted in the Italian context,  its generalizability could be questioned. However, the fact that the model trained on data coming from five cities was able to predict the remaining city's vitality levels speaks to the generalizability of our approach, at least in the Italian, if not European, context. \rebuttal{Rome posed a challenge though. Being ancient, Rome's city centre exhibits a unique mix of the old and the new, while our proxies for vitality are grounded on modern urban morphologies. A similar type of conflict between modern and ancient urban forms was also found for Barcelona \cite{delclos2018looking}, in which nearby districts tend to enjoy a highly diverse historical mix.} \rebuttal{For our models to better generalize to unseen cities, greater diversity in terms of urban, historical, cultural, and natural contexts in the training set might be required. } 




\rebuttal{Another limitation is that our models are primarily  capturing the \emph{potential} for - rather than actual - vitality, and, in certain - albeit rare - circumstances, their predictions might not match the actual vitality levels. For example, if we took the satellite images of Milan during the full COVID-19 epidemic lock-down (during which pedestrian activity was heavily limited), our models would still predict levels of vitality similar to those of ordinary times. This limitation is, however, shared with the majority of previous work, and could be partly fixed whenever higher-resolution satellite images (in which  pedestrian activity is observable) would be publicly available.}

 \rebuttal{Another line of future work is to integrate multi-modal data, given that vitality is, by definition, a multi-faceted concept. By fusing satellite data with other freely available datasets that, for example, capture mix of commercial activities and concentration of people, one is expected to improve prediction accuracy. Finally, to obtain explainable predictions without compromising  accuracy, explainable AI methods should be further researched \cite{samek2019towards}. }

\bibliographystyle{ACM-Reference-Format}
\bibliography{sample-base}

\appendix









\end{document}